\documentclass[runningheads]{llncs}
\usepackage{graphicx}
\usepackage{subcaption}
\usepackage{tikz}
\usepackage{array}
\usepackage{comment}
\usepackage{amsmath,amssymb} 
\usepackage{color}
\usepackage{enumitem}
\usepackage{appendix}
\usepackage{soul}
\usepackage{gensymb}
\newcolumntype{P}[1]{>{\centering\arraybackslash}p{#1}}
\usepackage[accsupp]{axessibility}  
\usepackage{multirow}
\usepackage{multicol}
\usepackage{authblk}
\usepackage{orcidlink}

\begin{document}
\pagestyle{headings}
\mainmatter

\title{Bokeh-Loss GAN: Multi-Stage Adversarial Training for Realistic Edge-Aware Bokeh} 
\titlerunning{Bokeh-Loss GAN for Realistic Edge-Aware Bokeh} 
\authorrunning{B. Lee et al.} 
\author{Brian Lee\inst{1}\orcidlink{0000-0002-6179-433X} \and
Fei Lei\inst{2}\orcidlink{0000-0002-4241-8769} \and
Huaijin Chen\inst{1}\orcidlink{0000-0002-7337-149X} \and
Alexis Baudron\inst{1}\orcidlink{0000-0002-6120-9778}}
%
%
\institute{Sensebrain Technology, San Jose CA 95131, USA \and
Tetras.AI, Shenzhen, China \\
\email{\{brianlee, chenhuaijin, alexis.baudron\}@sensebrain.site}\\
\email{leifei1@tetras.ai}}

\maketitle

\begin{abstract}
In this paper, we tackle the problem of monocular bokeh synthesis, where we attempt to render a shallow depth of field image from a single all-in-focus image. Unlike in DSLR cameras, this effect can not be captured directly in mobile cameras due to the physical constraints of the mobile aperture. We thus propose a network-based approach that is capable of rendering realistic monocular bokeh from single image inputs. To do this, we introduce three new edge-aware \textit{Bokeh Losses} based on a predicted monocular depth map, that sharpens the foreground edges while blurring the background. This model is then finetuned using an adversarial loss to generate a realistic Bokeh effect. Experimental results show that our approach is capable of generating a pleasing, natural Bokeh effect with sharp edges while handling complicated scenes. 

\end{abstract}

\section{Introduction}
    The Bokeh effect is a highly desirable aesthetic effect in photography used to make the subject stand out by blurring away the out-of-focus parts of the image. This effect can be achieved naturally in Single-lens Reflex (SLR) cameras by taking an image with a wide aperture lens and large focal length, producing an image with a shallow depth of field (DoF). 
    However, such shallow DoF effect can not be achieved naturally on mobile devices, due to the short focal lengths, small sensor and aperture sizes of mobile camera modules. As a result, the Bokeh effect can only be rendered synthetically on mobile devices, for example with stereo \cite{bokeh_stereo} or dual-pixel hardware \cite{synthetic_dof_mobile} that provides a disparity map. This task is further constrained when limited to a single, monocular camera which is common particularly on mid to lower-end mobile devices.

    In this paper, we propose a multi-stage, network-based approach that leverages both edge refinement, adversarial training and bokeh appearance loss to render a realistic bokeh effect from a single wide-DoF image. 
    As Figure \ref{fig:model} shows, our backbone bokeh generator network takes in a single image and its disparity map and outputs the rendered bokeh. We employ a depth estimation network to produce a monocular disparity map which we use as a blurring cue for our backbone bokeh network. We also utilize this disparity map as a grey-scale saliency mask, which is used to sharpen the foreground edges while smoothing the background through three new loss functions. These losses create a strong, yet slightly rough bokeh effect which is then refined adversarially through a dual-scale PatchGAN \cite{Pix2pix} discriminator that we train jointly with the model backbone in a process similar to \cite{bggan}. 


    \begin{figure}[h]
    \begin{center}
    \begin{subfigure}{.3\textwidth}
    \centering
    \includegraphics[width =.95\linewidth]{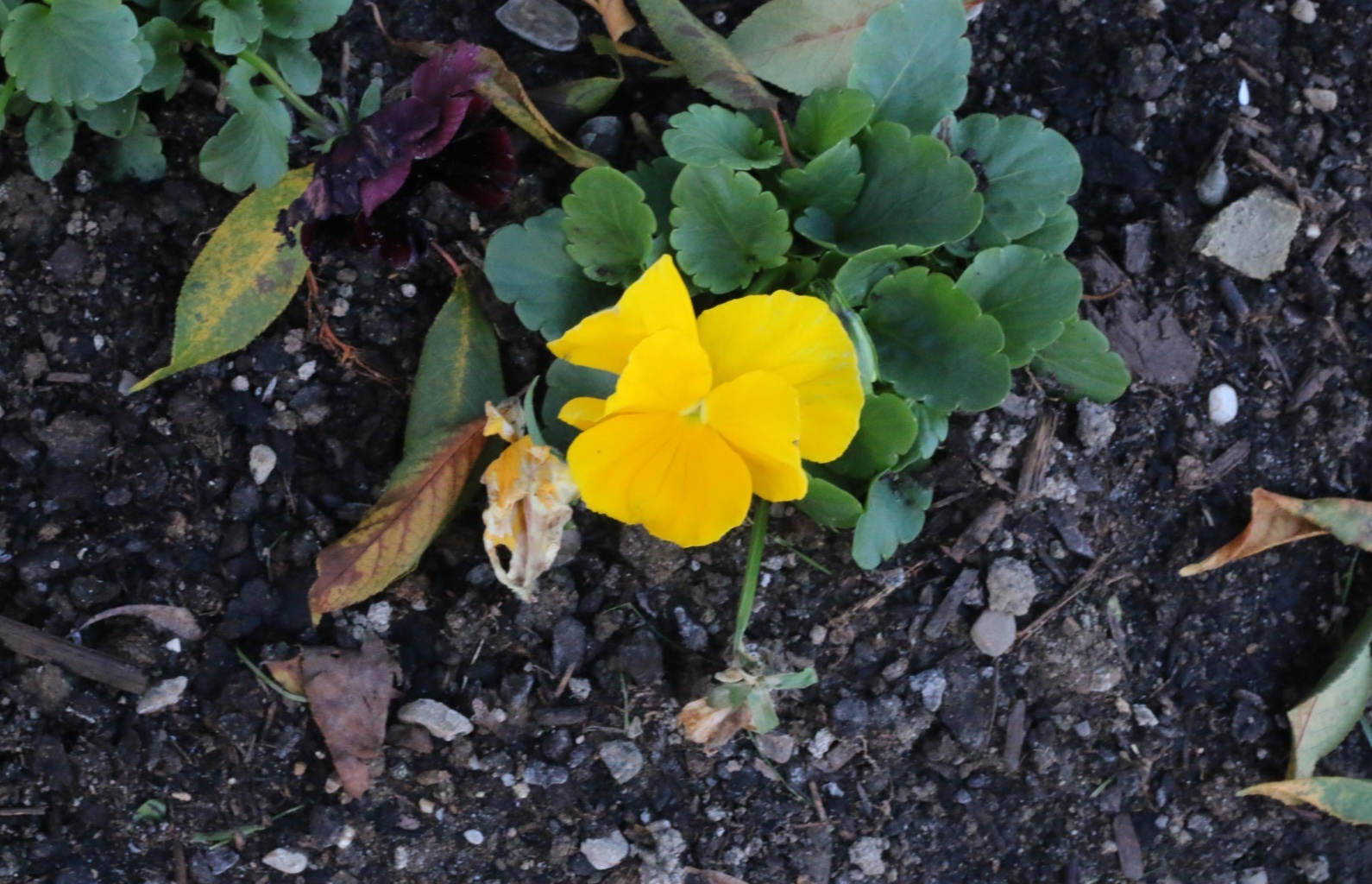}
    \subcaption{Original Image}
    \end{subfigure}
    \begin{subfigure}{.3\textwidth}
    \centering
    \includegraphics[width = .95\linewidth]{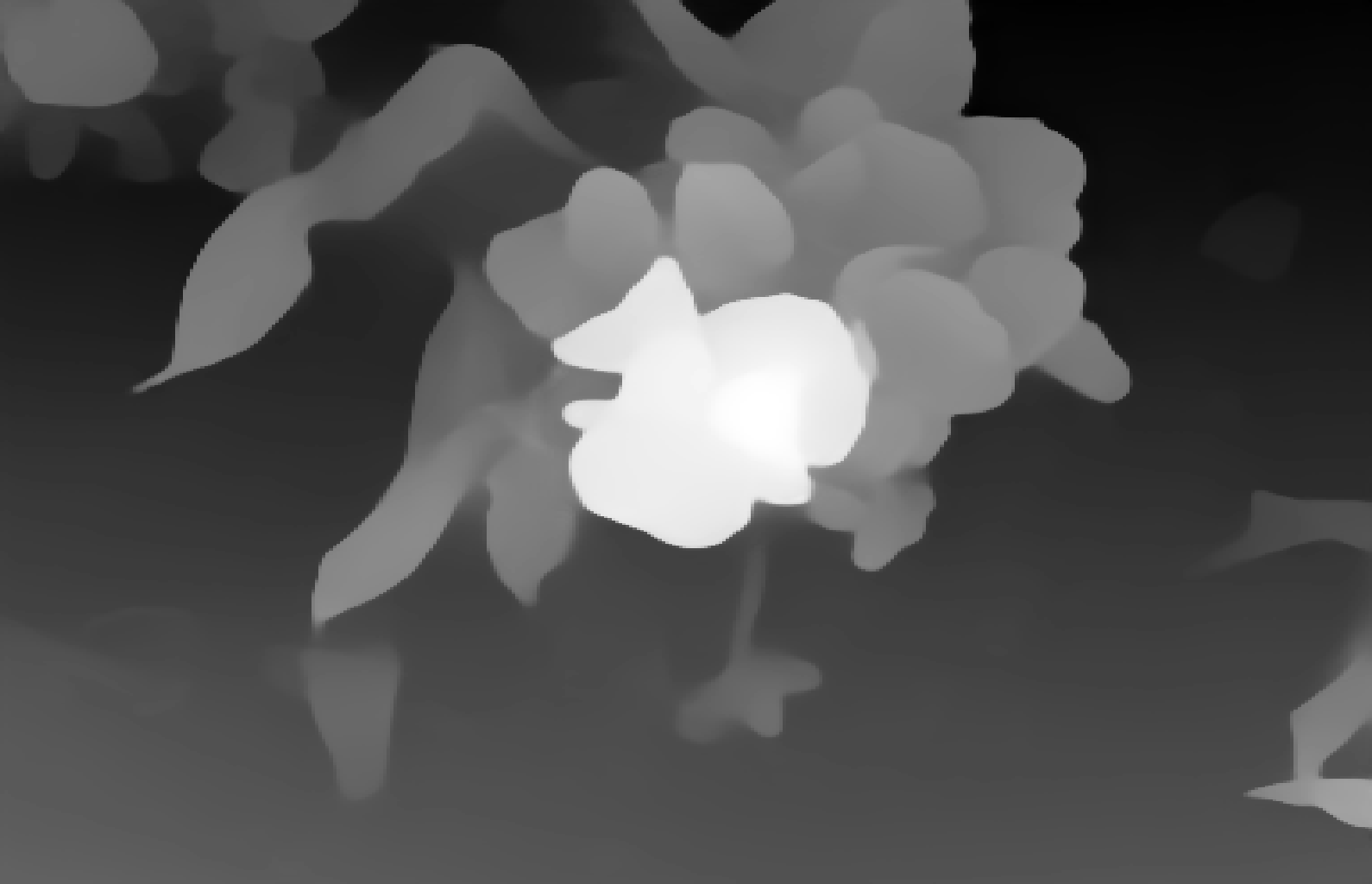}
    \subcaption{Disparity Map}
    \end{subfigure}
    \begin{subfigure}{.3\textwidth}
    \centering
    \includegraphics[width = .95\linewidth]{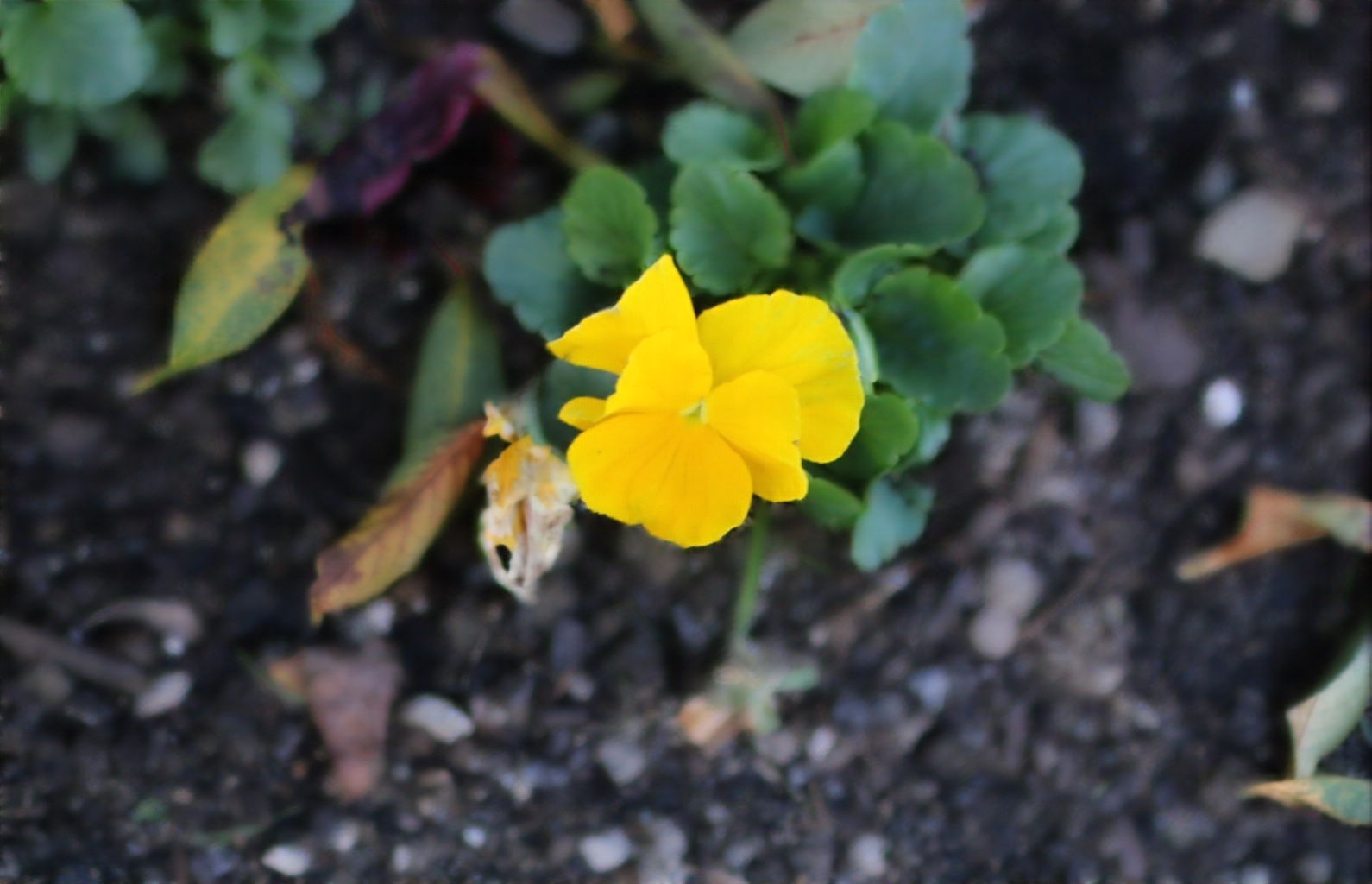}
    \subcaption{Bokeh Output}
    \end{subfigure}
    \end{center}
    \label{intro_ex}
    \caption{Given a single wide depth-of-field image, our model estimates the corresponding disparity map using a Depth Prediction Transformer (DPT). This map is then concatenated with the input image and then run through the trained NAFNet model to produce a shallow depth-of-field image. Best viewed from computer screen.}
    \end{figure}

    In summary, our main contributions are:
    \begin{enumerate}[label=(\arabic*)]
        \item Three new loss functions that use a greyscale saliency mask for edge-aware bokeh rendering from monocular images taken with wide depth-of-field
        \item A multi-stage training scheme that adversarially refines the output produced using the three aforementioned losses
        \item A computationally efficient blurring backbone, successfully applying the NAFNet architecture to the task of Bokeh
    \end{enumerate}

\section{Related Work}

\subsection{Computational Bokeh}

A variety of different methods have been explored for synthetic Bokeh rendering in recent years. Classical rendering methods take into account the physics of the image scene, before combining various different modules to construct an automatic Bokeh render. Earlier methods, such as \cite{auto_portrait_segmentation,deep_portrait_matting,fast_deep_matting} were limited to images in portrait mode. These methods relied on the segmentation of the image foreground from the background, before applying a filter on the background to achieve the Bokeh effect. Another class of methods, such as in \cite{sterefo,highquality_uncalibrated,interactive_portrait_bokeh,deeplens,ranking-based-sod,3dreconstruction}, blurred the image by making use of predicted depth maps, either through depth estimates from monocular depth module \cite{megadepth,DPT,Midas}, stereo vision \cite{fast_bilateral_stereo,improved_dm_estimation}, or through a moving camera using the parallax effect \cite{depth_motion_parallax,joint_parallax}. The input image is then decomposed to multiple layers conditioned on the estimated depth map, and then rendered back to front to prevent bleeding into the foreground. As for the actual blur, many different methods have been adopted based on imaging physics including convolutional filters such as in \cite{auto_portrait_segmentation}, or through physics-based scattering methods such as in \cite{interactive_portrait_bokeh,synthetic_dof_mobile}. 

Recent advancements in Bokeh rendering have also seen the adoption of neural-network based rendering methods to avoid boundary artifacts. For example, \cite{deep_shading,deepfocus} train a neural network to predict Bokeh effects from perfect (synthetic) depth maps. Real perfect depth maps are hard to come by, however, so other end-to-end network models have been introduced, such as in \cite{stacked_deep_multiscale,pynet,aim2020,bggan}. These models all use some sort of encoder-decoder structure to process the image at multiples scales and use approaches such as monocular depth estimation or image segmentation to improve bokeh results. Recent work by Peng \textit{et al.} \cite{Peng2022BokehMe} has also suggested combining the classical and network-based approaches to improve controllability in the Bokeh process. 

\subsection{Monocular Depth Estimation}

Many different approaches have been proposed for monocular depth estimation. Earlier work by Eigen \textit{et al.} \cite{eigen_depth} proposed a multi-scale DNN based on AlexNet to generate a depth estimate. This method, however, suffered a number of key dataset limitations, namely a small number of training samples and a lack of variety. 

Li \textit{et al.} \cite{megadepth} proposed a new depth data collection system that allowed for better generalization of depth models. This was followed by a breakthrough by Ranftl \textit{et al.} \cite{Midas}, who proposed a method for mixing multiple depth datasets even with incomplete annotations. Both of these methods used a convolutional architectures to create the depth prediction. Ranftl \textit{et al.} \cite{DPT} then showed that a transformer backbone could be used to further improve the depth prediction quality. 
\subsection{Generative Models}
Generative Models \cite{gan_original} have been shown to generate realistic images while preserving fine texture details. These models, however, have been shown to often suffer from the problem of Modal Collapse. To remedy this, algorithms such as WGAN \cite{wasserstein_gan} and WGAN-GP \cite{wgan-gp} have been proposed to enhance training stability.

Recent work has shown the success of GAN-based architectures on image translation tasks such as in image deblurring \cite{deblurgan}, single-image super resolution \cite{srgan,real-esrgan,esrgan}, semantic segmentation \cite{cell_image_segment}, and so on. Conditional Adversarial Networks \cite{Pix2pix}, in particular, have shown to generalize to arbitrary image translation tasks. Recently, this framework has also been adopted to the task of monocular bokeh synthesis \cite{bggan} in refining the bokeh rendering quality.  

\section{Proposed Method}\label{method}

Rendering an accurate bokeh effect from an all-in-focus image is a complicated task that requires processing the image at multiple scales. For a network to learn an accurate blur effect, it must (a) learn how to accurately segment out the in-focus object regions by processing the image at a global scale and (b) learn an accurate blur on the out-of-focus regions by processing the image on a local scale. When the image is poorly segmented (a), however, blur in the out-of-focus regions (b) can often bleed into the foreground, creating an unnatural bokeh effect at boundary edges. Tackling both tasks (a) and (b) in a single end-to-end neural network is thus extremely challenging without additional cues that aid in the individual tasks. 

\begin{center}
\begin{figure}[h]
\begin{center}
\includegraphics[width=\textwidth]{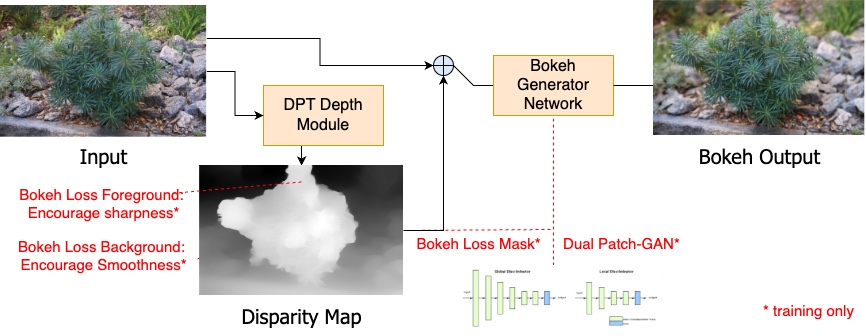}
\end{center}
\caption{Bokeh-Loss GAN pipeline: We first generate the depth of the scene from the all-sharp input, then combine the depth map and the input image, and feed them to the bokeh generator network to produce the bokeh image. Besides conventional L1 and SSIM losses, we introduce a depth-weighted bokeh loss, in addition to a dual-scale discriminator loss.}
\label{fig:model}
\end{figure}
\end{center}

We add a series of modules that aid in both tasks during training and inference. As shown in Figure \ref{fig:model}, the major components of our network are:
\begin{enumerate}[label=\roman*.]
    \item A Dense Prediction Transformer (DPT) \cite{DPT} based depth estimation module that estimate the relative depth of the scene.
    \item A Nonlinear Activation Free Network (NAFNet) based generator module \cite{nafnet} that takes in the all-in-focus image and depth map of the scene, and outputs the depth-aware bokeh image.
    \item A dual-receptive field patch-GAN discriminator similar to the one proposed in \cite{bggan}.
    \item A bokeh model based loss function that encourages sharp content in the in-focus area and smooth content in the background area.
\end{enumerate}

The bokeh output is generated by first running the input image through the depth prediction module to produce a depth map $\textbf{D}$. The output $\textbf{D}$ is then concatenated with the original input image $\textbf{I}$ and then run through the NAFNet generator $\textbf{G}$ to produce a shallow depth-of-field output $\textbf{G}(\textbf{I} \odot \textbf{D})$. 

\subsection{Backbone Network}\label{NAFNet}

Given the translational nature of the Bokeh task, we adopt a backbone architecture that performs well on such problems. Various backbone networks have been proposed recently for image restoration tasks, such as \cite{nafnet,Restormer}. In particular, the NAFNet architecture \cite{nafnet} has been shown to maintain a relatively small memory footprint while achieving exceptional output performance on such tasks. This is achieved by replacing high complexity non-linear operators such as Softmax, GELU, and Sigmoid with simple multiplications, thereby reducing the complexity in computationally expensive attention estimation.


Due to the similar nature of the prediction task, we therefore use the NAFNet architecture for our Bokeh backbone. Experimental results have shown (see section \ref{Nafnet_size_ablation}) that on the scale necessary for mobile devices (around width $12$ or less), that the difference in NAFNet width does not result in that large of a difference in the quality of the output images. We thus use a smaller NAFNet model with a width of $8$. We empirically set the number of NAFblocks for the encoder to be [2,2,2,20] and the number of NAFBlocks for the decoder. We also concatenate 2 NAF blocks in the middle for a total of $36$ NAFBlocks in the network. The total number of parameters comes to around 1.047M parameters.

\subsection{Depth Estimation Network}

Our model uses a depth map estimate during both training and inference stages as a blur and edge cue (see section \ref{bokeh_loss}). The depth map of the scenes are estimated from a single monocular image by a dense prediction transformer (DPT) \cite{DPT}. We use the depth prediction model that was pre-trained on a large-scale mixed RGB-to-depth ``MIX 5'' datasets \cite{Midas}. We find empirically that such networks can generate realistic depth maps with smooth boundaries and fine-grain details at object boundaries. 

\begin{figure}[h]
\begin{center}
\begin{subfigure}{.32\textwidth}
    \centering
    \includegraphics[width =.93\linewidth]{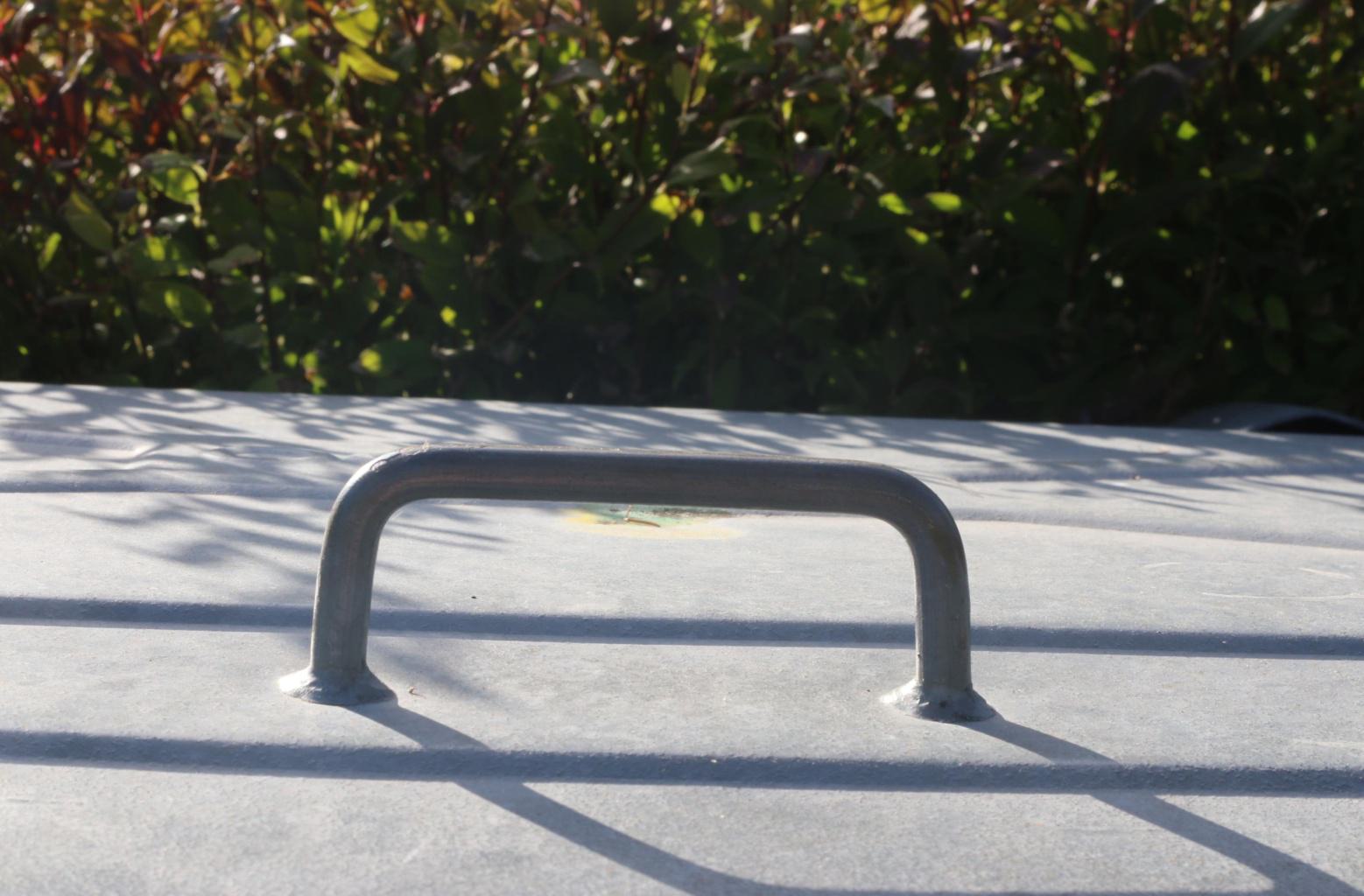}
\end{subfigure}
\begin{subfigure}{.32\textwidth}
    \centering
    \includegraphics[width = .93\linewidth]{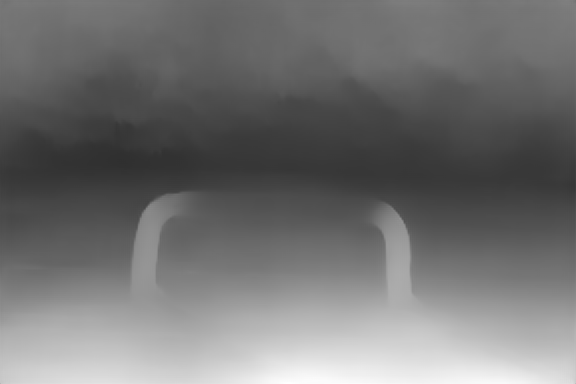}
\end{subfigure}
\begin{subfigure}{.32\textwidth}
    \centering
    \includegraphics[width = .94\linewidth]{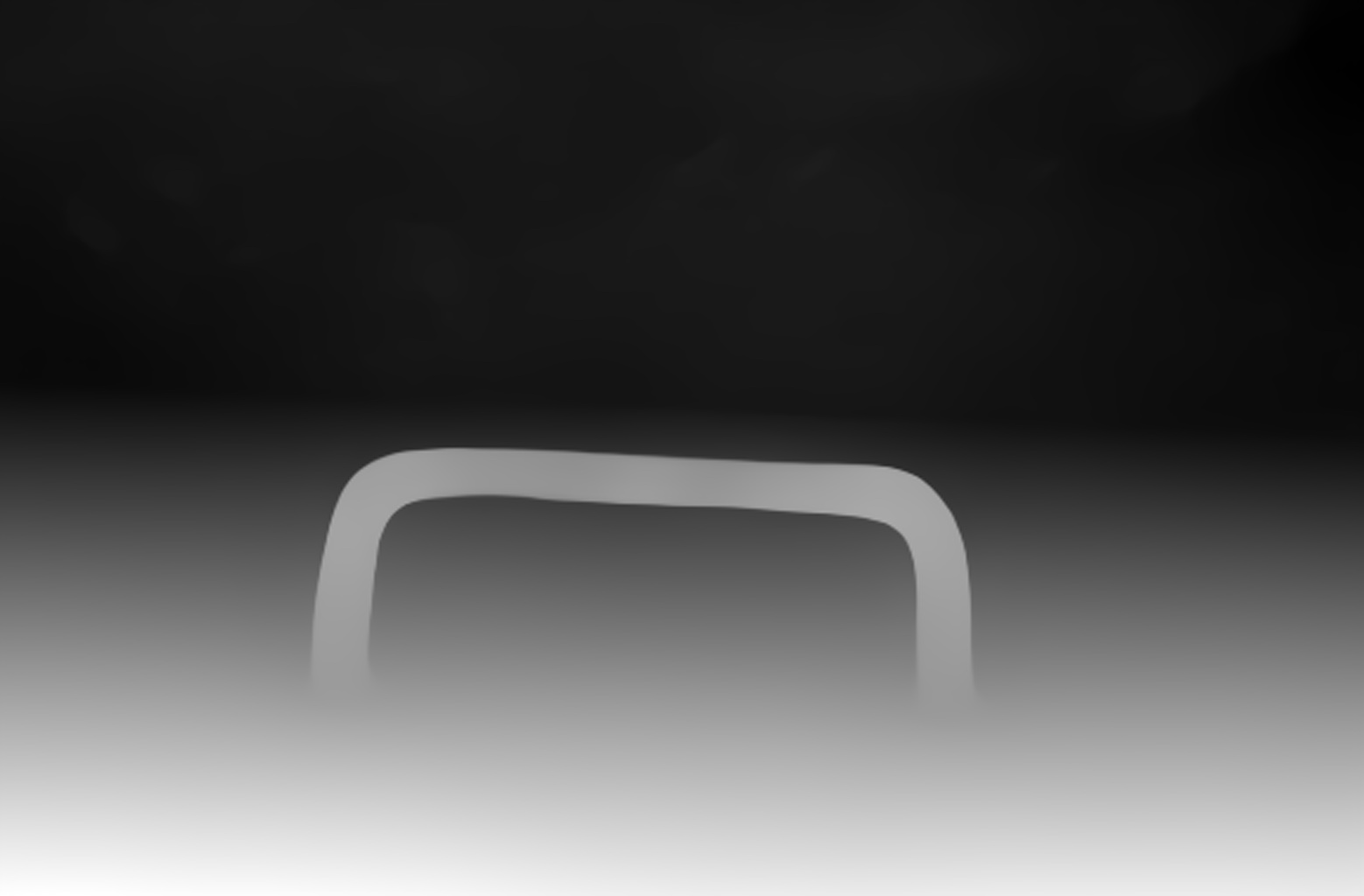}
\end{subfigure}
\end{center}
\caption{Comparison of Input images (left) with Depth Maps generated with MegaDepth (center) and DPT (right) for two EBB inputs}
\end{figure}

The generated depth map provides extra blurring clues to assist the backbone generator in creating better bokeh blur. Particularly, we found that accurate depth boundaries help to generate noticeably more natural looking bokeh boundaries than in previous methods. 


\subsection{Discriminator Loss}\label{discriminator_loss}
To improve perceptual quality of the final Bokeh output, we utilize a discriminator loss to ensure the perceived visual quality of the generated bokeh images. Similar to \cite{bggan}, we use a multi-receptive-field Patch-GAN discriminator \cite{Pix2pix} as part of our loss function. 
\begin{figure}[h]
\begin{center}
\includegraphics[scale = 0.22]{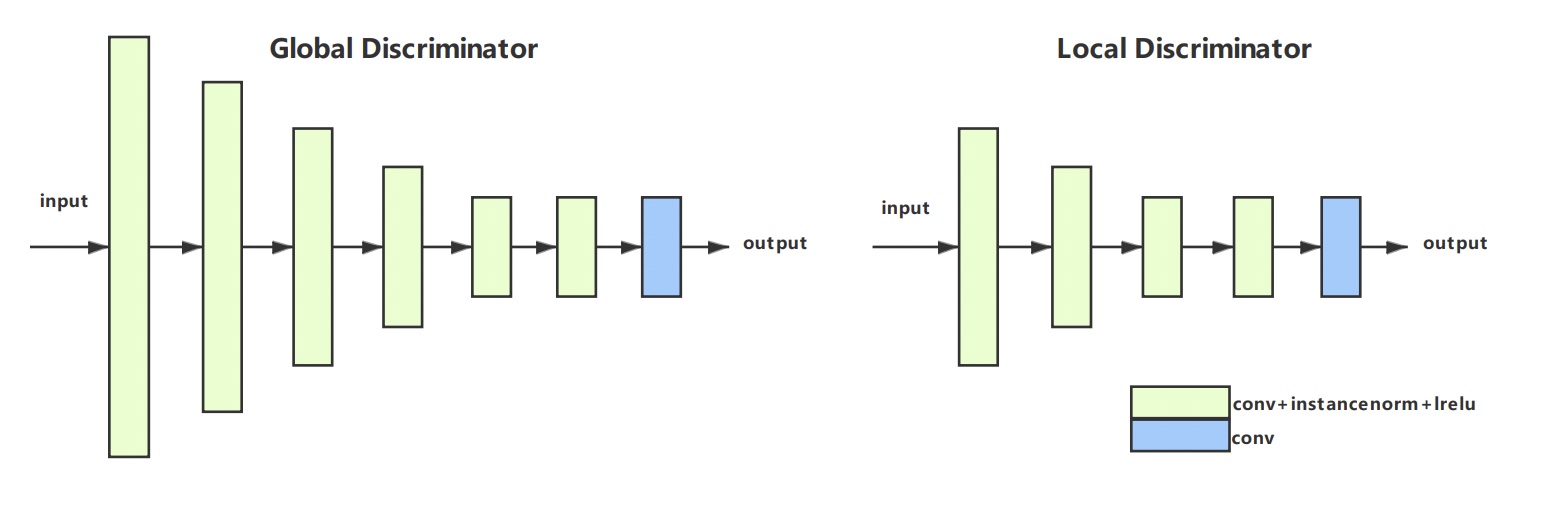}
\caption{Multi Receptive Discriminator}
\end{center}
\end{figure}
The model takes two patch-GAN discriminators, one of depth $3$ and one of depth $5$ and averages the adversarial losses from both discriminators into one loss. We use a $\lambda$ value of $1$ for the WGAN gradient penalty for the discriminator loss.

\subsection{Bokeh Loss}\label{bokeh_loss}
In addition to the conventional L1, SSIM, and discriminator loss functions, we propose a bokeh-specific loss. The core idea is that, for a natural bokeh, the in-focus area should be sharp and the out-of-focus area should be blurred. Using the depth map we generate in the DPT module as a greyscale saliency mask $\textbf{M}$, we introduce the following loss functions:  

First, we introduce a \textit{Foreground Edge Loss} that tries to maximize the intensity of the foreground edges. To do this, we first multiply our input image $\hat{\textbf{I}}$ with the greyscale saliency mask $\textbf{M}$ to obtain the in-focus regions of the image. We compute the edge map (gradients) using Sobel filters in the $0\degree$, $90\degree$ and $\pm45\degree$ directions which provides richer structure information, we then sum up the L1-norms of the gradients, before normalizing. We then take the negative of this sum to maximize the foreground edge intensity rather than minimize it. The loss is computed as follows: 
\begin{align}
L_{foreedge}(\hat{\textbf{I}}, \textbf{M}) = - \frac{\sum_{z \in \{0\degree, \pm45, \degree90\degree\}}\left\|S_{z}(\hat{\textbf{I}}\cdot \textbf{M})\right\|_1}{h_{\hat{\textbf{{I}}}}\cdot w_{\hat{\textbf{I}}}}
\end{align}
where $S_z$ is the sobel convolution operator in the $z$ direction to maximize the intensities of the foreground.

While we wish to strengthen the intensity of the edges however, we also want to create edges that are of a similar intensity to that of the output image. To achieve this, we add an \textit{Edge Difference Loss}:
\begin{align}
L_{edgediff}(\textbf{I}, \hat{\textbf{I}}, \textbf{M}) = ||L_{foreedge}(\hat{\textbf{I}}\cdot\textbf{M})|-|L_{foreedge}(\textbf{I} \cdot \textbf{M})||
\end{align}
that attempts to minimize the difference in foreground edge strength between the input and the output image. 

Finally, we add a \textit{Background Blur Loss} $L_{backblur}$ that encourages a smoother blur for the background. We first multiply the input image $\hat{\textbf{I}}$ with the inverse of the greyscale saliency mask $\textbf{M}$ to obtain the out-of-focus regions. We then try to minimize the total variation of the scene:
\begin{align}
L_{backblur}(\hat{\textbf{I}}, \textbf{M}) = \frac{1}{h_{\hat{\textbf{I}}}\cdot w_{\hat{\textbf{I}}}}TV(\hat{\textbf{I}} \cdot (1-\textbf{M}))
\end{align}
where the added factor is used to normalize the loss. 

In summary, we add three new loss functions:
\begin{itemize}
\item A \textit{Foreground Edge Loss} that encourages sharper edges seperating the foreground and the background
\item An \textit{Edge Difference Loss} that encourages similar edge intensities for the input and output image
\item A \textit{Background Blur Loss} that encourages a smoother background with less noise.
\end{itemize}

Qualitative results have shown that these losses induce sharp, albeit slightly distorted edges, along with a smooth background, which when combined with the adversarial loss, appears to increase the overall MOS of the output (see section \ref{bokehloss_ablation}).

\subsection{Training Setup}

In order to get the sharp edges and smooth background induced by the Bokeh losses while maintaining a natural blur and avoiding edge artifacts, we adopt a dual-stage strategy for model training. 

In the first, ``pretraining'' stage, we use a weighted sum of L1 loss, SSIM loss, and our three aforementioned ``Bokeh'' losses, using the depth map as a greyscale mask to separate the foreground from the background. The final loss-function for this stage was given by 
\begin{align}
L_{pretrain} = 0.5 \times L_{1} &+ 0.05 \times L_{SSIM} + 0.005 \times L_{edgediff} \\
&+ 0.1 \times L_{backblur} + 0.005 \times L_{foreedge} \nonumber
\end{align}
where the final weights were determined experimentally. This pretraining stage then creates a rough bokeh effect with sharp, albeit slightly distorted edges and a smooth background. 

In the second stage, the model is finetuned with a weighted sum of L1 loss, SSIM loss, VGG perceptual loss, and an adversarial loss from the dual-scale PatchGAN discriminator that is trained jointly in the second stage using the WGGAN-GP method \cite{wgan-gp}. This second stage refines the distorted edges and the background to generate a realistic bokeh effect. The refinement loss thus took the form 
\begin{align}
L_{refinement} = 0.5 \times L_{1} &+ 0.1 \times L_{VGG} + 0.05 \times L_{SSIM} + L_{adv}
\end{align}
where the weights were again determined experimentally. The adversarial loss $L_{adv}$ is weighted the highest with a coefficient of $1$ in order to allow the discriminator to play the leading role in the refining phase. 

\section{Experiments}

In this section, we introduce our experimental setting and compare both the qualitative and quantitative performance of our solution to current state-of-the-art architectures that were designed for this problem, along with other models that were submitted as part of the AIM 2022 \textit{Real-Time Rendering Realistic Bokeh} challenge. We also conduct a series of ablation studies to analyze the different effects of various factors. 

\subsection{Experimental Setup}

\subsubsection{Technical Specifications}
Our method uses standard Pytorch packages and the Pytorch Lightning framework for training. All models were trained on $8$ $32$G Nvidia V$100$ GPUs, $383$G RAM, and $40$ CPUs. 

In training, we cropped all images to $1024 \times 1024$ sized patches as inputs. The final models were all trained with batch size $2$, with $60$ epochs for the first stage and $60$ epochs for the second stage. For optimization, we use the Adam optimizer with learning rate $1e-4$ for both the backbone and the discrminator, with $(\beta_1,\beta_2)=(0,0.9)$. 

\subsubsection{Dataset}
The \textit{Everything is Better with Bokeh!} (EBB!) dataset \cite{pynet} was released as part of the AIM 2022 \textit{Real-Time Rendering Realistic Bokeh Challenge}. The dataset consists of $4696$ shallow/wide depth-of-field image pairs for training and $200$ images for validation and testing respectively. The image pairs were shot with a Cannon 7D DSLR camera taken with a narrow aperture (f/16) for the all-in-focus input image and a wide aperture (f/1.8) for the Bokeh ground truth image. The photos are taken in a wide variety of scenes, all in automatic mode, and are aligned using SIFT keypoint matching and the RANSAC method as in \cite{ransac_method}, before being cropped to a common region. Despite this, we found that many of the images suffered from either poor alignment or inconsistent lighting. We thus manually pruned the training dataset down to $4425$ images, to help the model learn a better bokeh effect. We then took out $200$ images for validation (which we call \textit{Val200}) before submitting our final results.

\subsection{Quantitative and Qualitative Evaluation}
\subsubsection*{Quantitative Evaluation}
Our model was submitted to the AIM 2022 \textit{Real-Time Rendering Realistic Bokeh Challenge} \cite{ignatov2022bokeh}, where the goal is to achieve shallow depth-of-field with the best perceptual quality compared to the ground truth .



Our solution was submitted only to Track 2 (unconstrained time) due to flex delegates that were used in the conversion from Pytorch to TFLite which makes our model unable to run on a mobile GPU. Our model achieved the best Learned Perceptual Image Patch Similarity  (LPIPS) \cite{lpips} score (lower is better), as shown in Table \ref{table: quantitative_results} which shows the performance of our model. Our model also yielded high scores on fidelity metrics, achieving the second best PSNR and SSIM scores. All models were run on a Kirin 9000 5G Processor, with the corresponding runtime being checked through the AI Benchmark \cite{aibenchmark} app. 

\begin{table}[h!]
\begin{center}
\begin{tabular}{| P{2.4cm} | P{1.5cm} | P{1.5cm} | P{1.5cm} | P{1.5cm} | P{2.4cm} |}
\hline
Team & PSNR $\uparrow$ & SSIM $\uparrow$ & LPIPS $\downarrow$ & Avg. Runtime(s) \\
\hline
xiaokaoji & 22.76 & 0.8652 & 0.2693 & \textbf{0.125} \\
\hline
MinsuKwon & \textbf{22.89} & \textbf{0.8754} & 0.2464 & 1.637 \\
\hline
hxin & 20.08 & 0.7209 & 0.4349 & 1.346 \\
\hline
sensebrain & 22.81 & 0.8653 & \textbf{0.2207} & 12.879 \\
\hline 

\end{tabular}
\end{center}
\caption{Quantitative results of our method from the Unconstrained CPU track of AIM 2020 \textit{Rendering Realistic Bokeh} Challenge.}
\label{table: quantitative_results}
\end{table}

Note that the runtime shown in the table above is the runtime of the TFLite model. The runtime of the pytorch model on CPU is $0.95$s. 
\subsubsection*{Qualitative Evaluation}

\begin{figure}[h!]
    \begin{center}
    \begin{subfigure}{.24\textwidth}
    \centering
    \includegraphics[width = \linewidth]{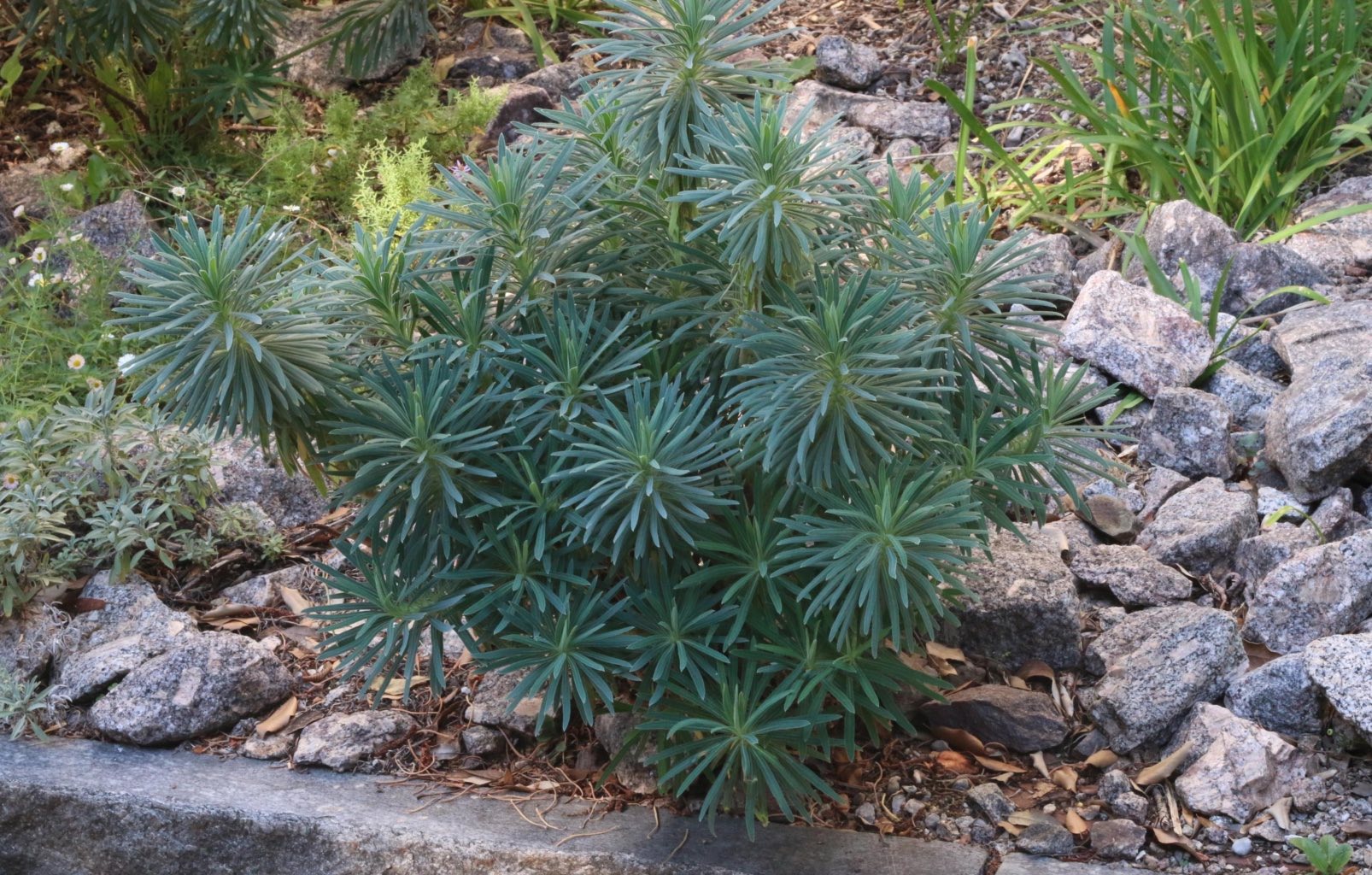}
    \end{subfigure}
    \begin{subfigure}{.24\textwidth}
    \centering
    \includegraphics[width = \linewidth]{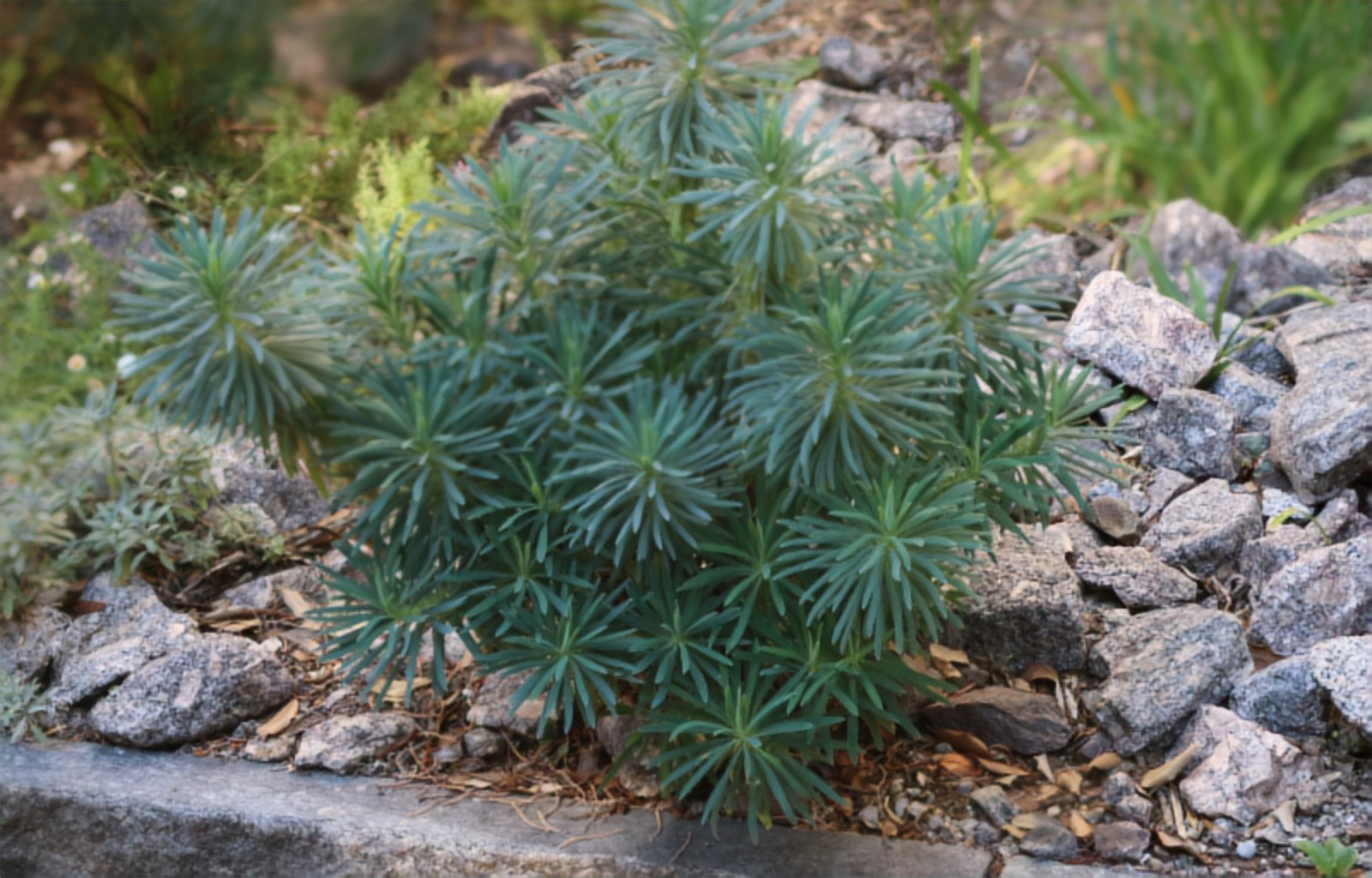}
    \end{subfigure}
    \begin{subfigure}{.24\textwidth}
    \centering
    \includegraphics[width = \linewidth]{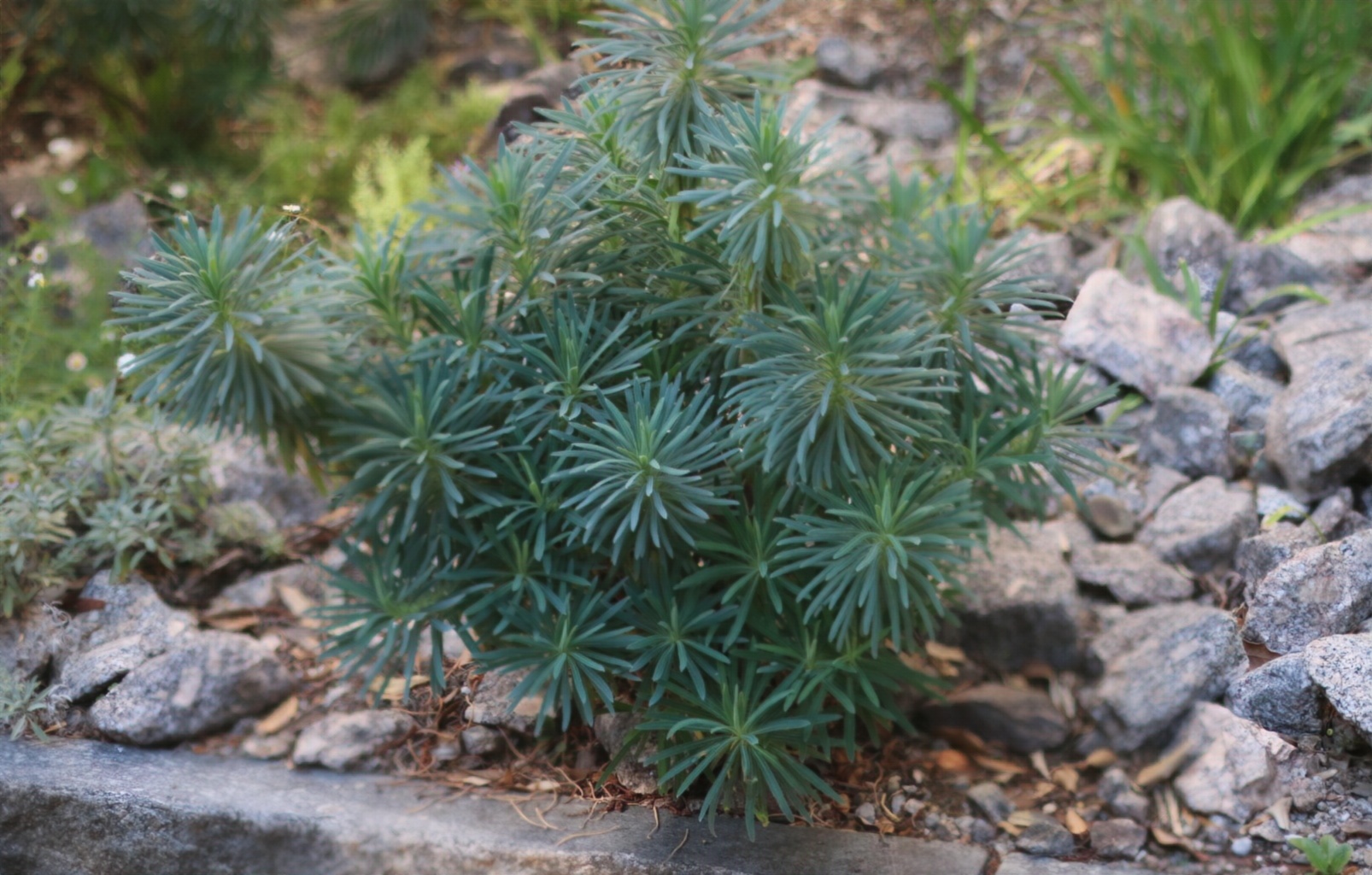}
    \end{subfigure}
    \begin{subfigure}{.24\textwidth}
    \centering
    \includegraphics[width = \linewidth]{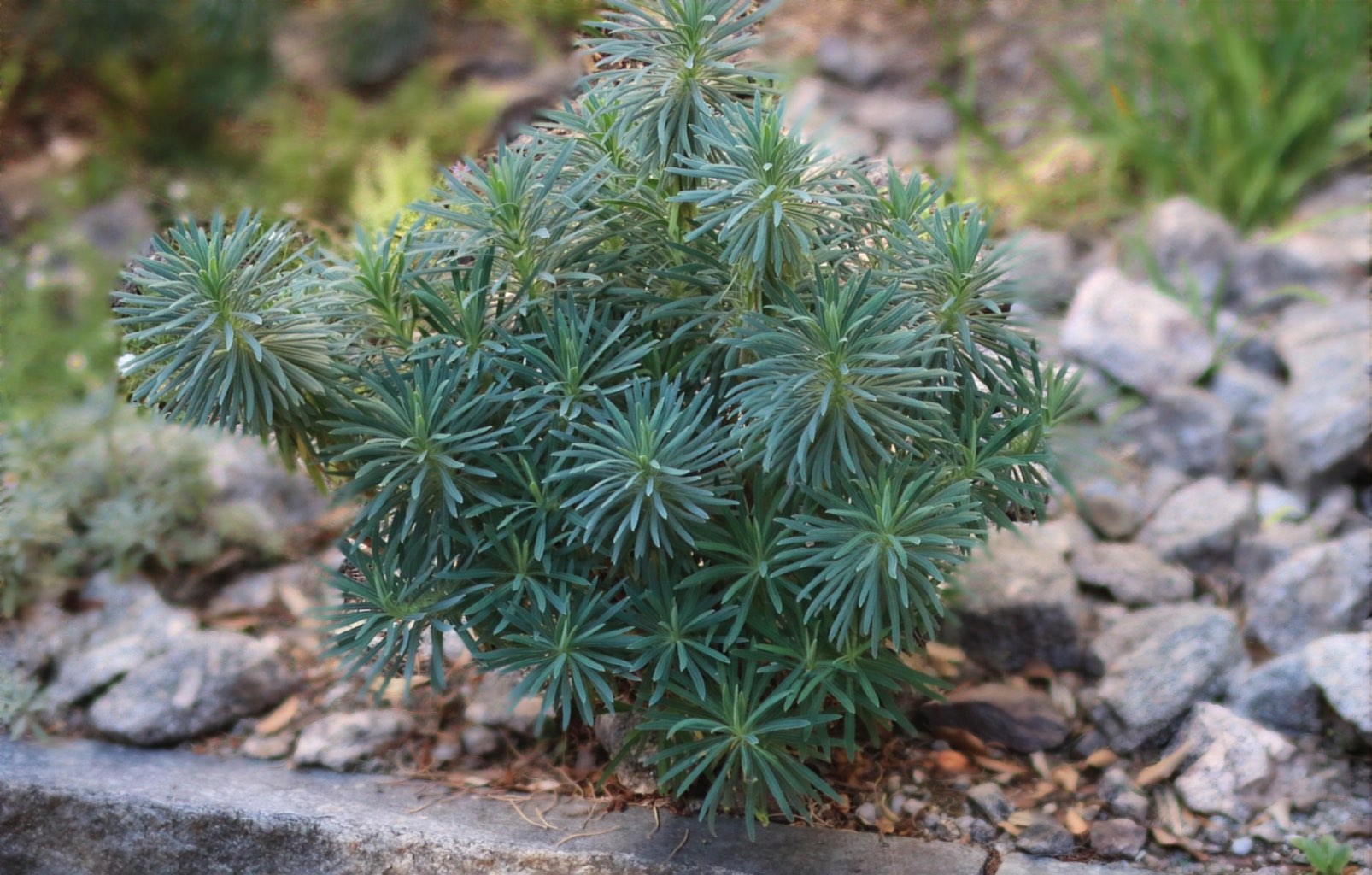}
    \end{subfigure} \\
    \begin{subfigure}{.24\textwidth}
    \centering
    \includegraphics[width = \linewidth]{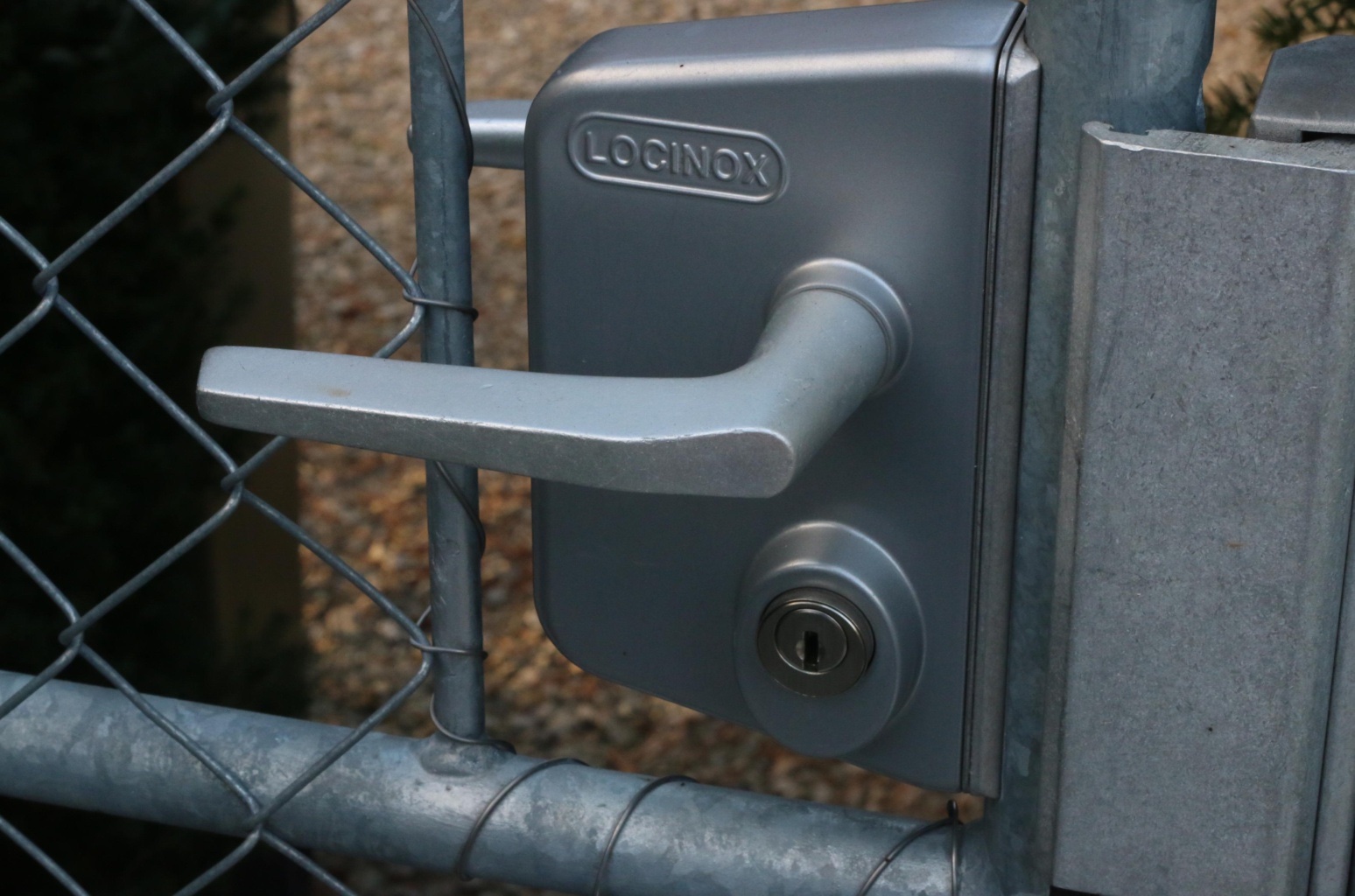}
    \end{subfigure}
    \begin{subfigure}{.24\textwidth}
    \centering
    \includegraphics[width = \linewidth]{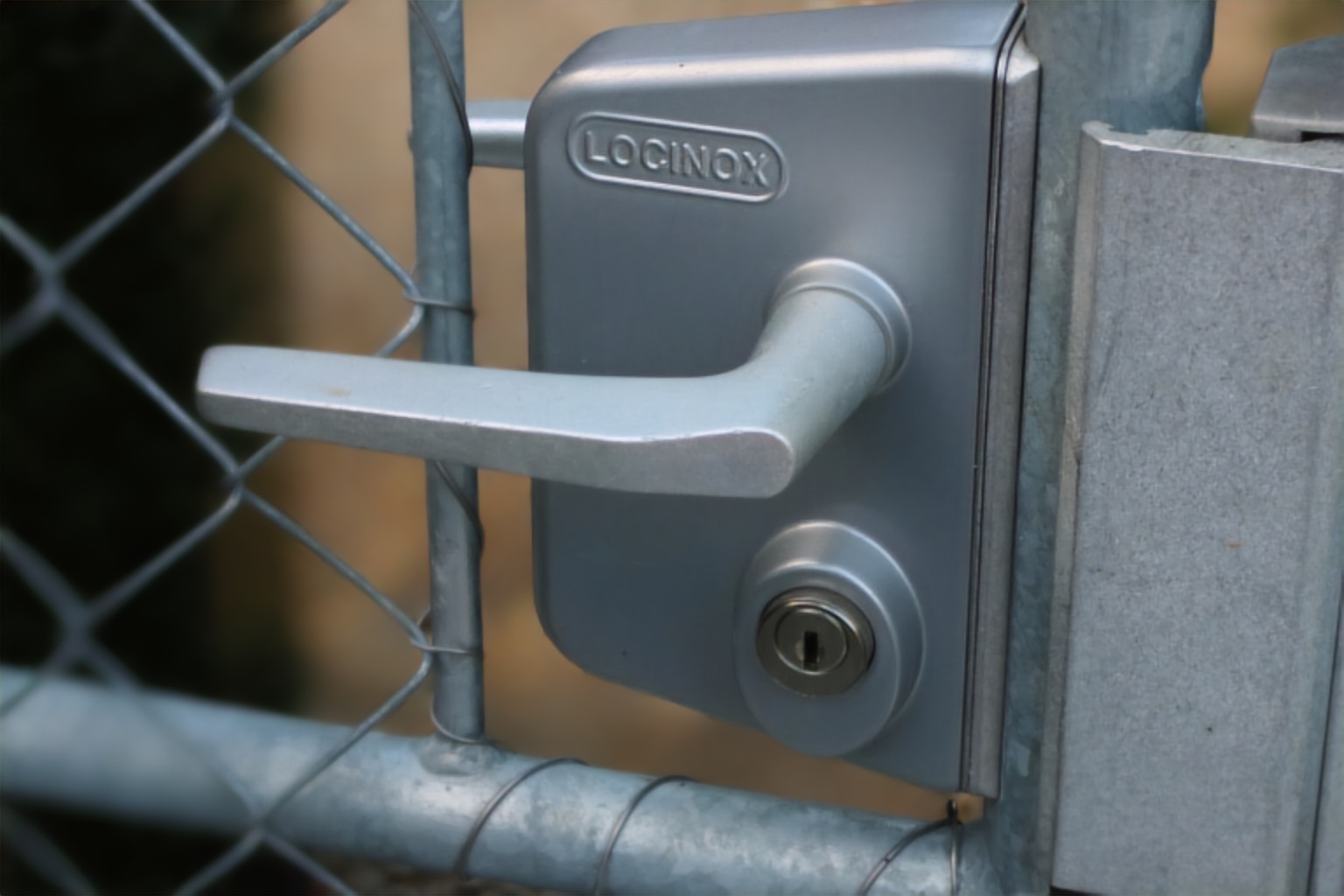}
    \end{subfigure}
    \begin{subfigure}{.24\textwidth}
    \centering
    \includegraphics[width = \linewidth]{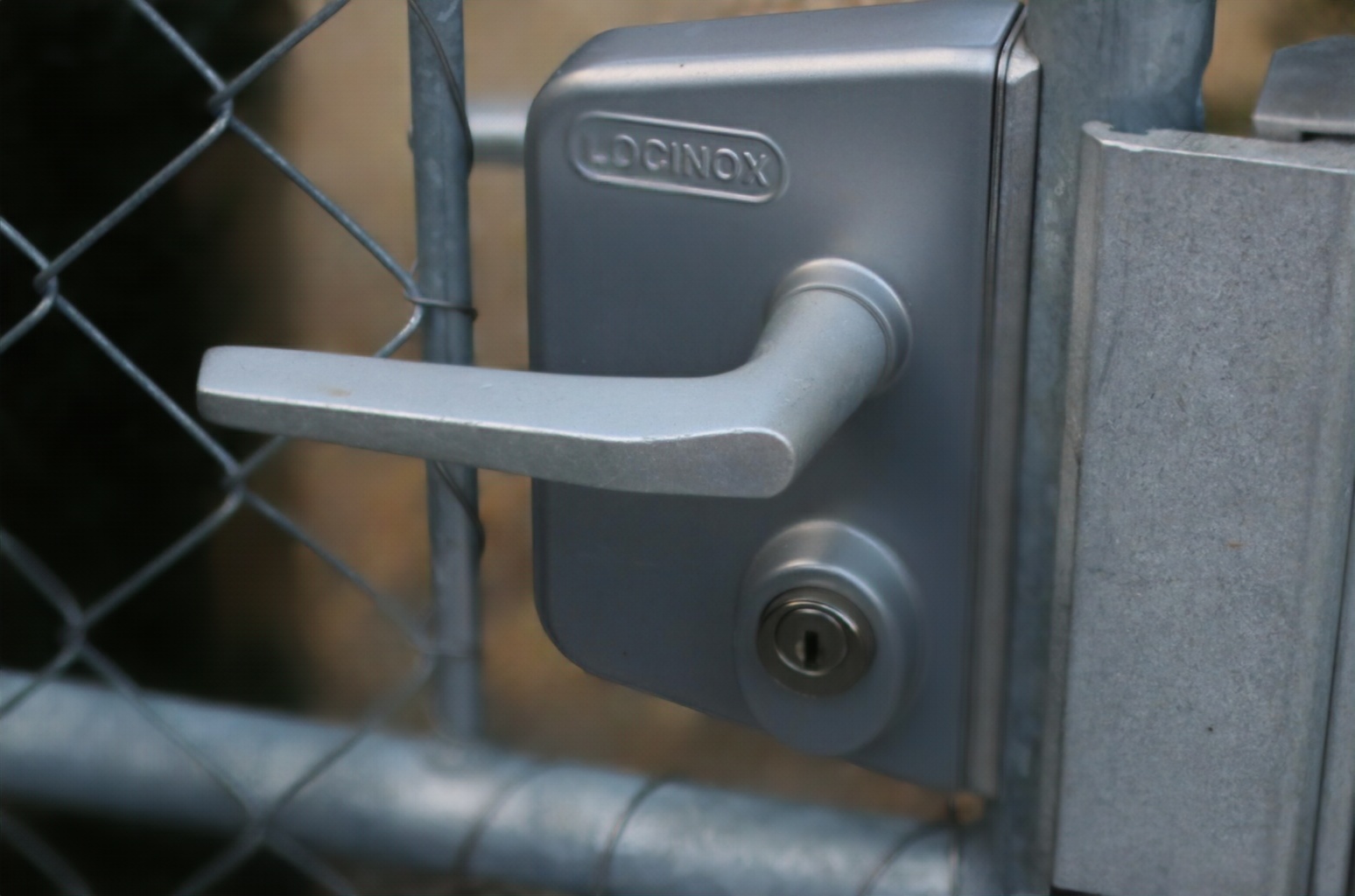}
    \end{subfigure}
    \begin{subfigure}{.24\textwidth}
    \centering
    \includegraphics[width = \linewidth]{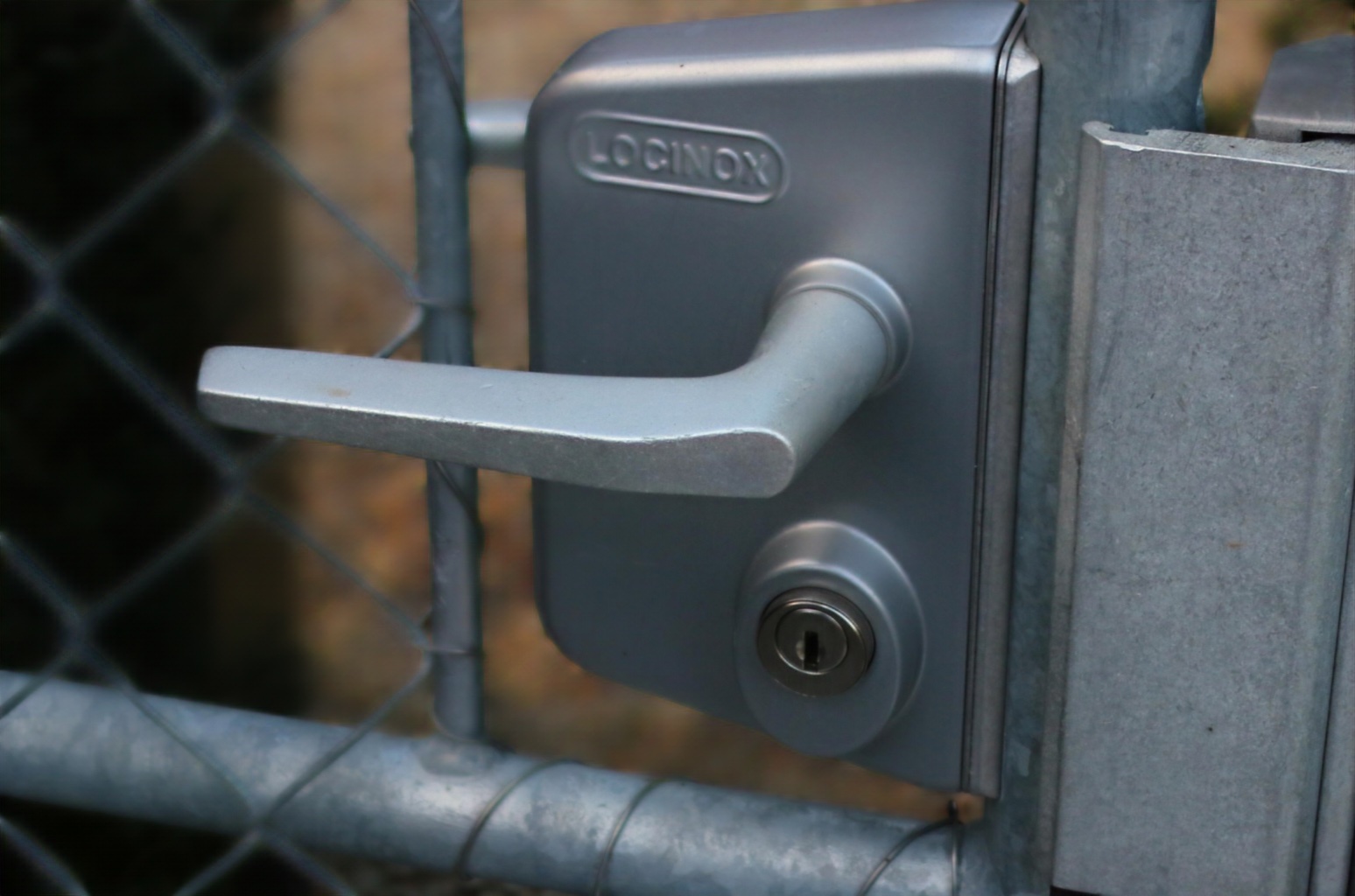}
    \end{subfigure} \\
    \begin{subfigure}{.24\textwidth}
    \centering
    \includegraphics[width = \linewidth]{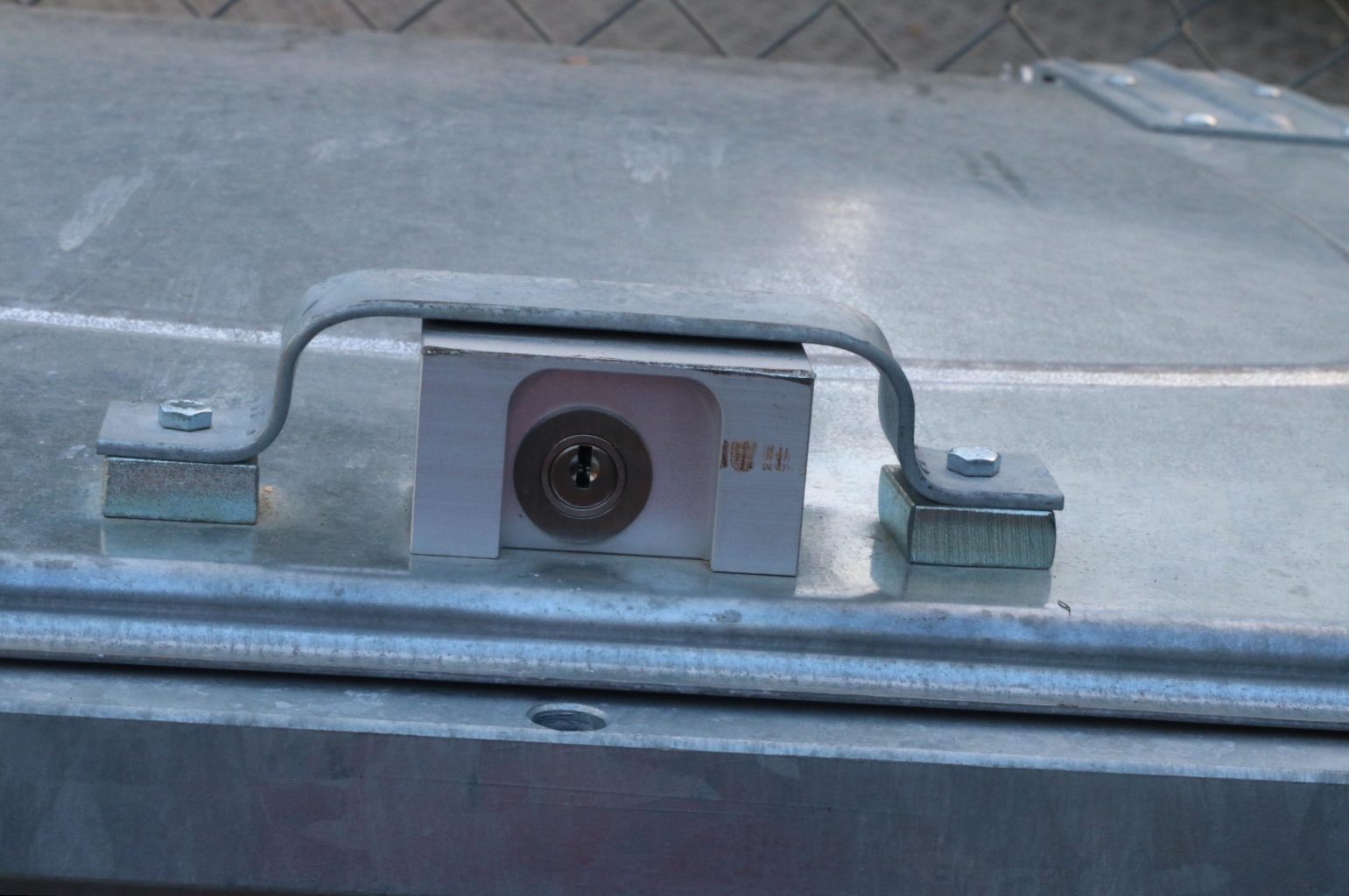}
    \end{subfigure}
    \begin{subfigure}{.24\textwidth}
    \centering
    \includegraphics[width = \linewidth]{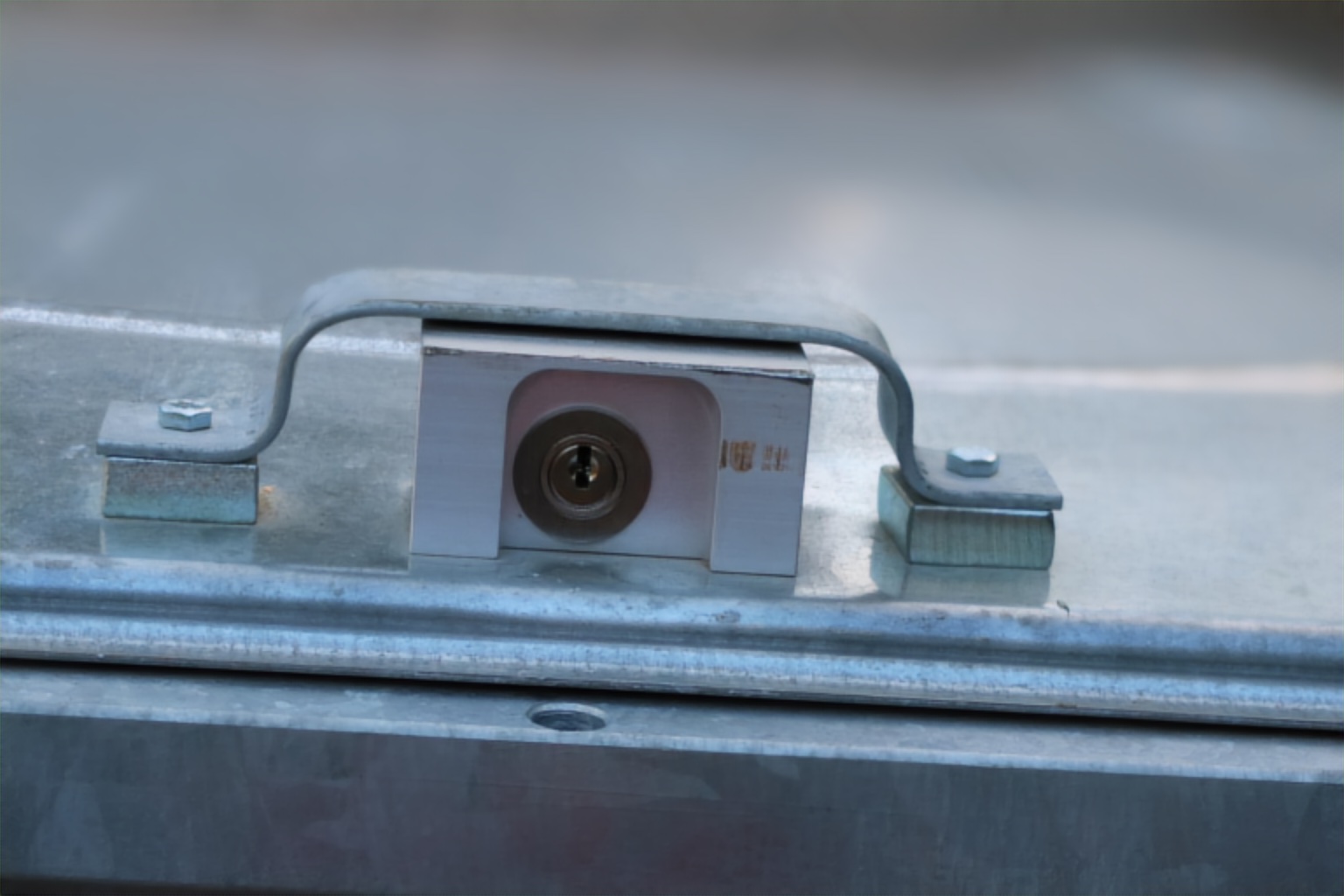}
    \end{subfigure}
    \begin{subfigure}{.24\textwidth}
    \centering
    \includegraphics[width = \linewidth]{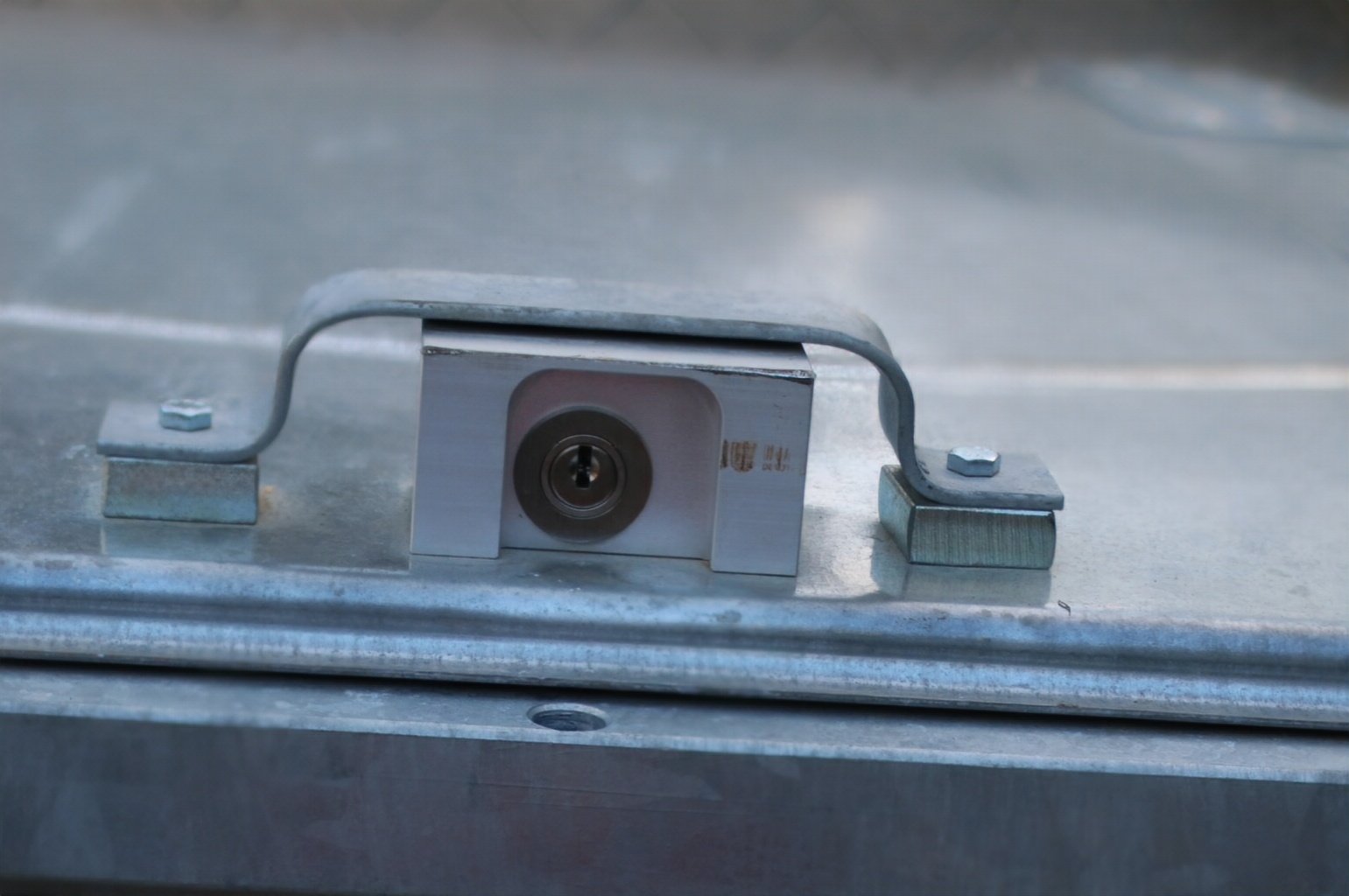}
    \end{subfigure}
    \begin{subfigure}{.24\textwidth}
    \centering
    \includegraphics[width = \linewidth]{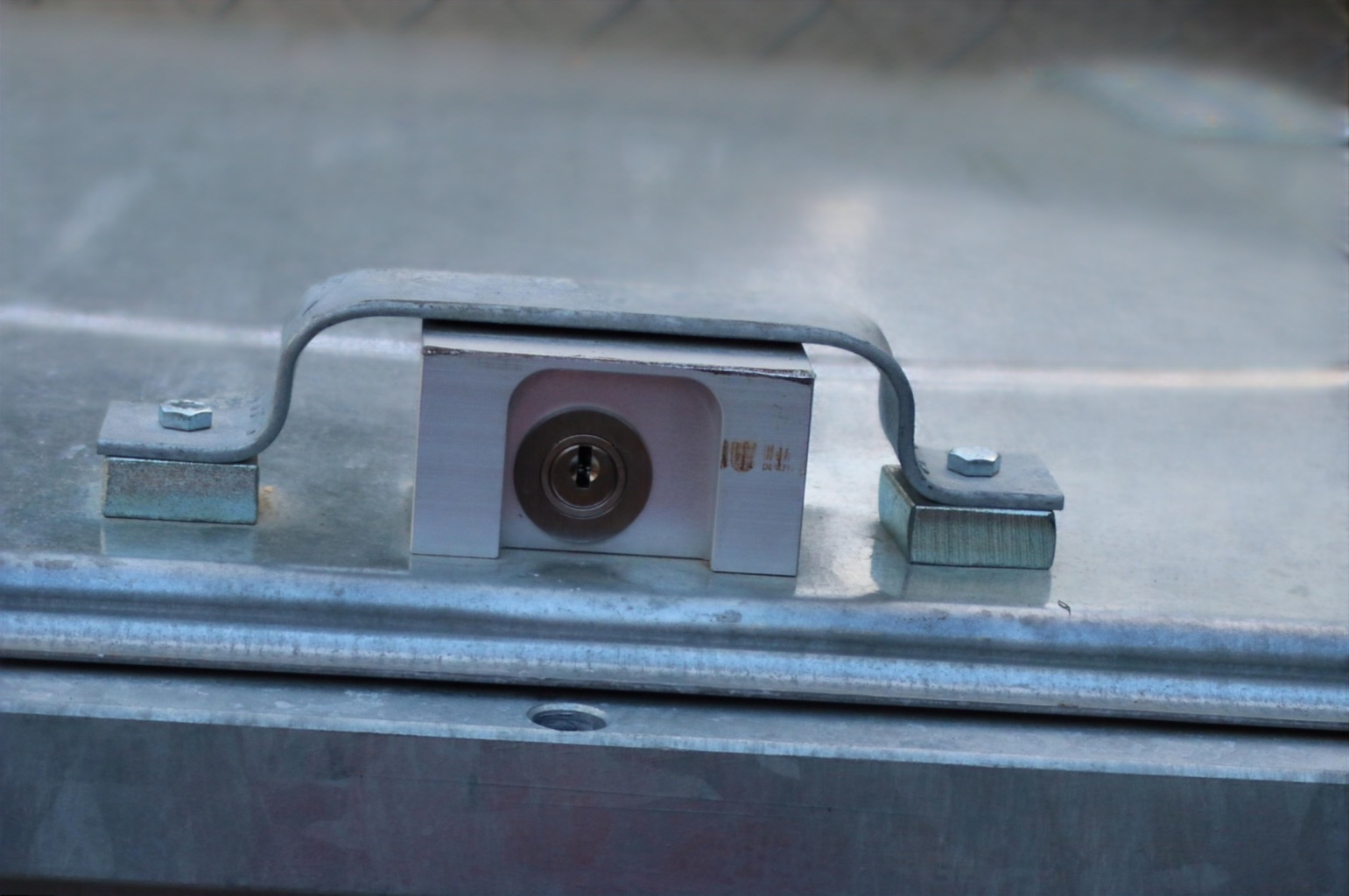}
    \end{subfigure} \\
    \begin{subfigure}{.24\textwidth}
    \centering
    \includegraphics[width = \linewidth]{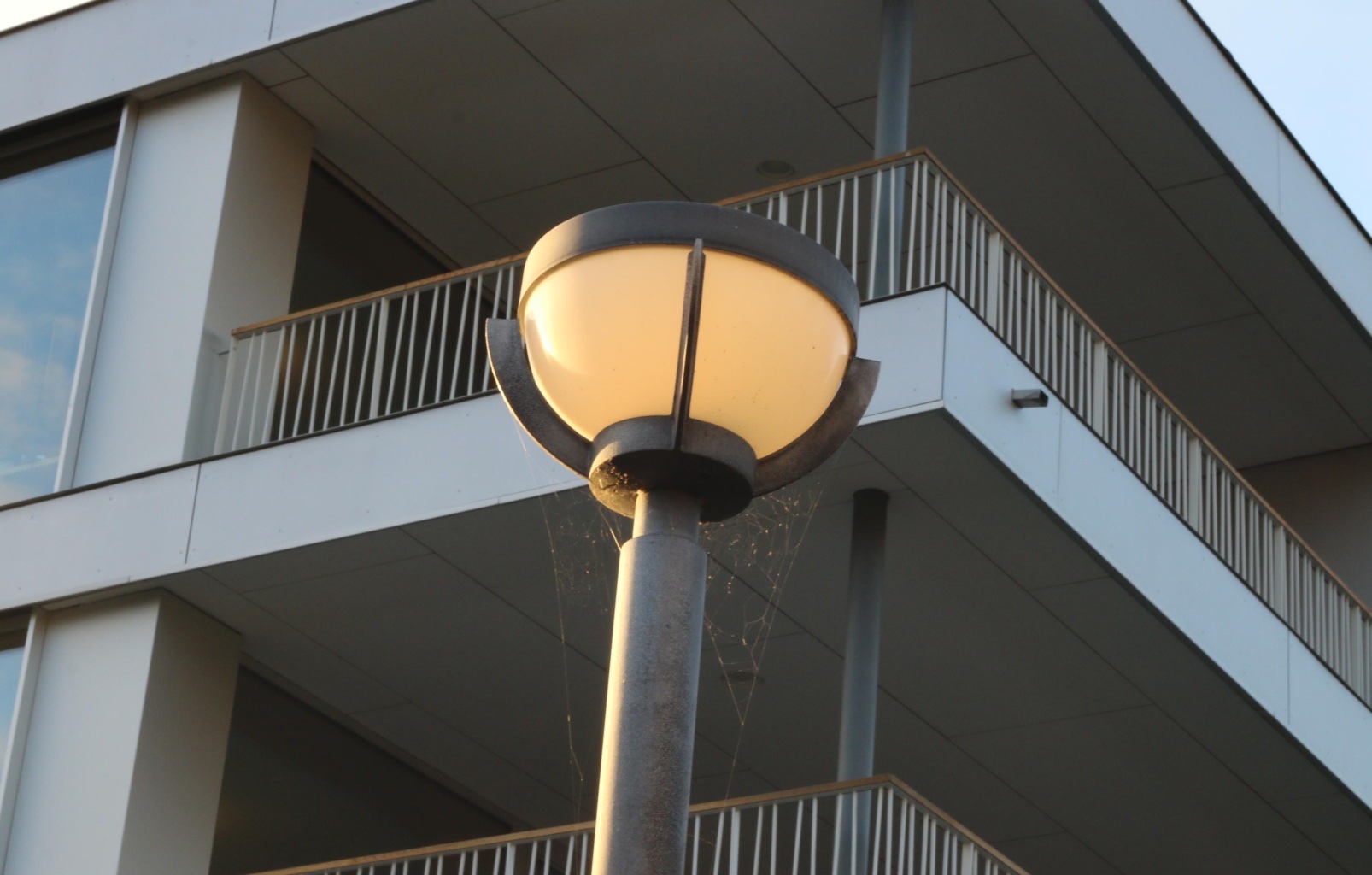}
    \end{subfigure}
    \begin{subfigure}{.24\textwidth}
    \centering
    \includegraphics[width = \linewidth]{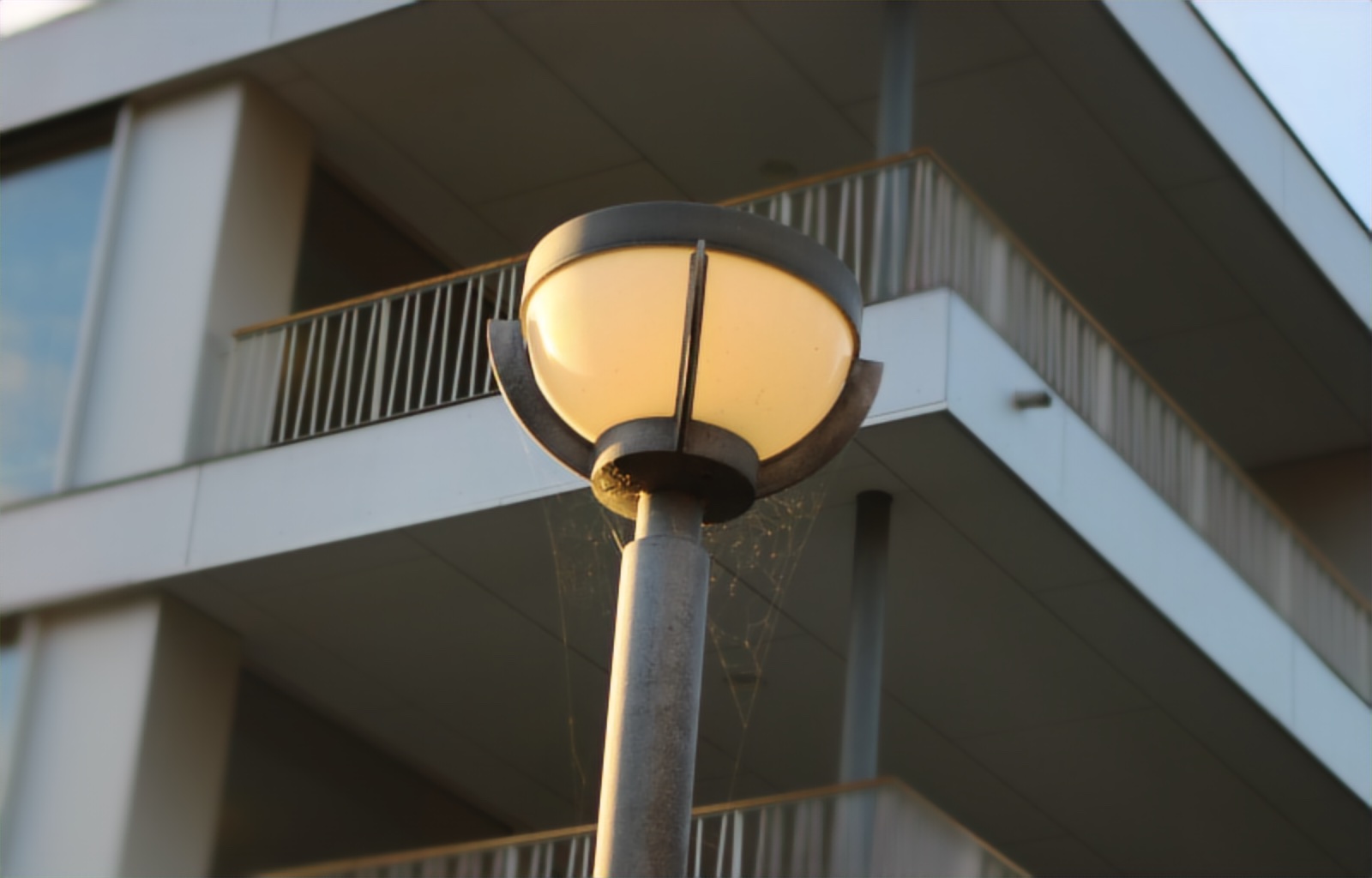}
    \end{subfigure}
    \begin{subfigure}{.24\textwidth}
    \centering
    \includegraphics[width = \linewidth]{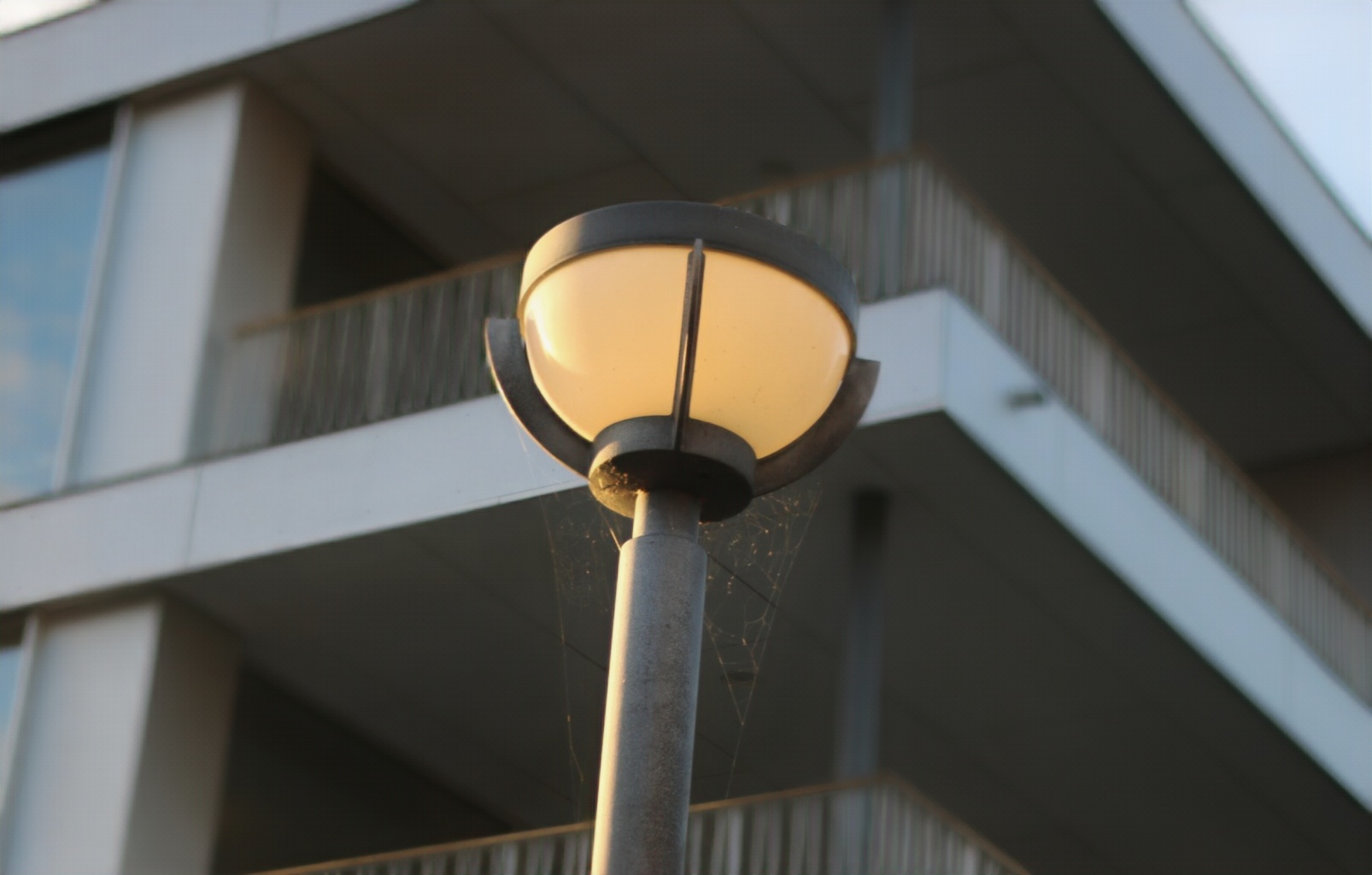}
    \end{subfigure}
    \begin{subfigure}{.24\textwidth}
    \centering
    \includegraphics[width = \linewidth]{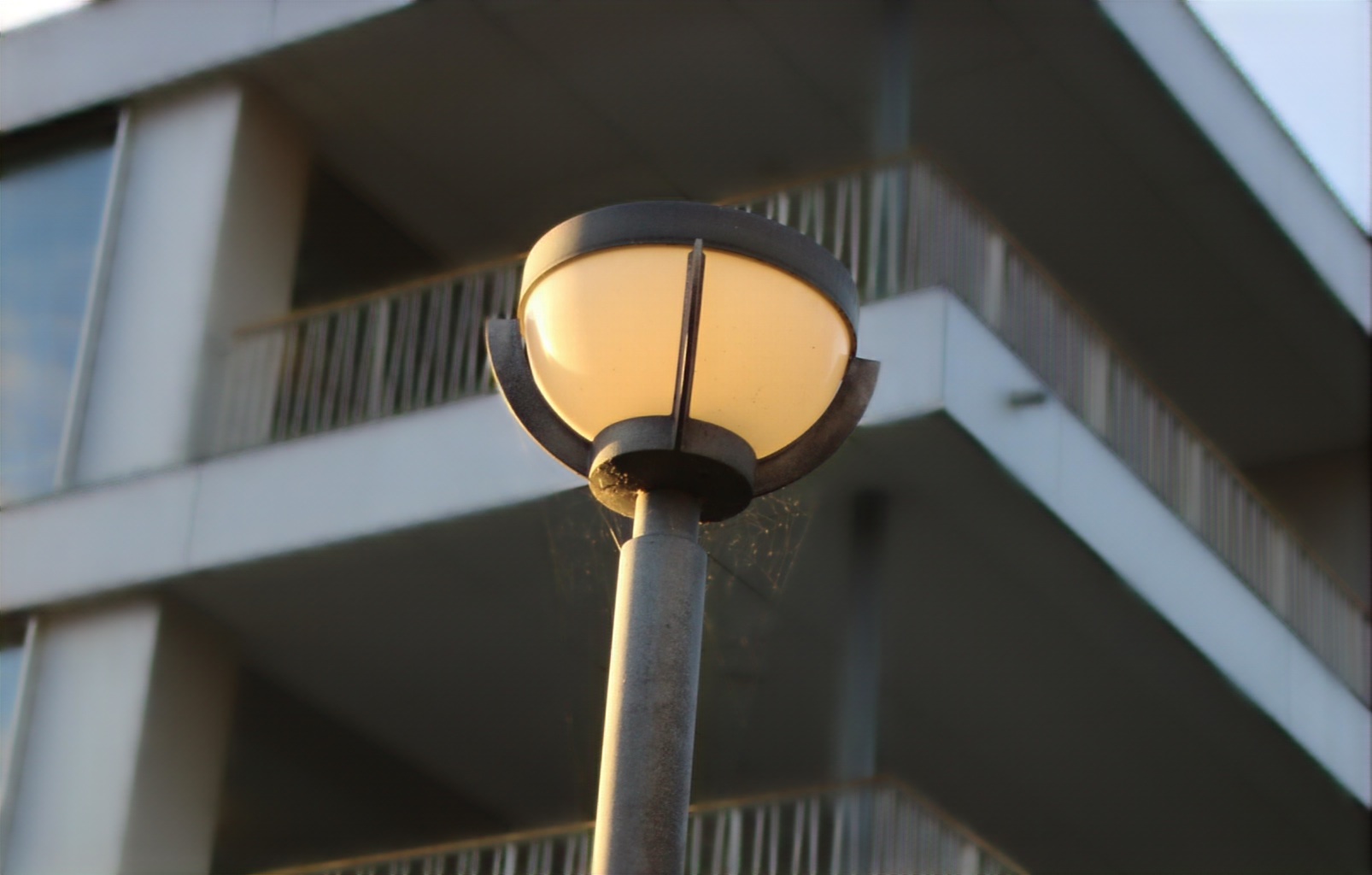}
    \end{subfigure} \\
    \begin{subfigure}{.24\textwidth}
    \centering
    \includegraphics[width = \linewidth]{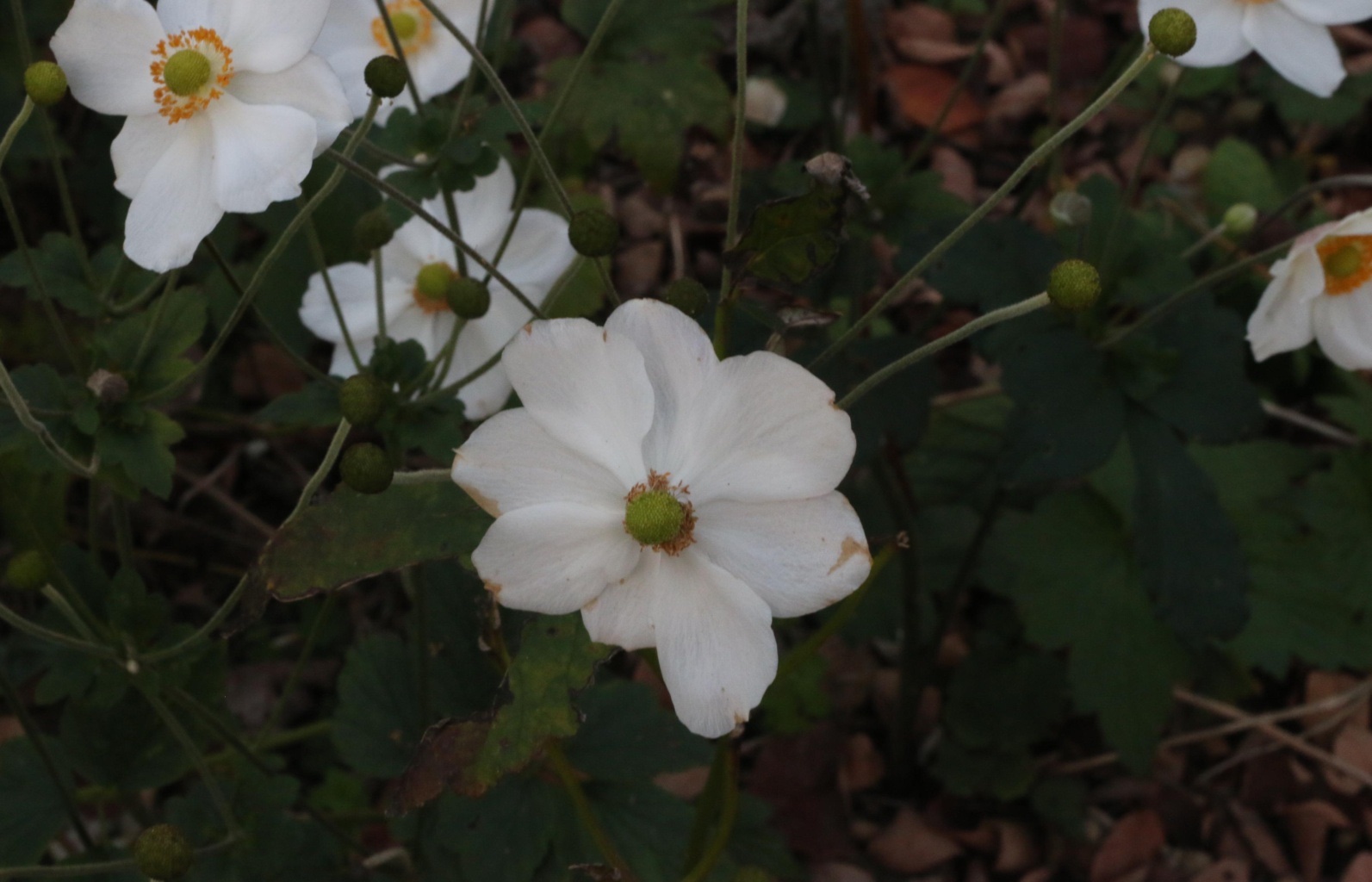}
    \end{subfigure}
    \begin{subfigure}{.24\textwidth}
    \centering
    \includegraphics[width = \linewidth]{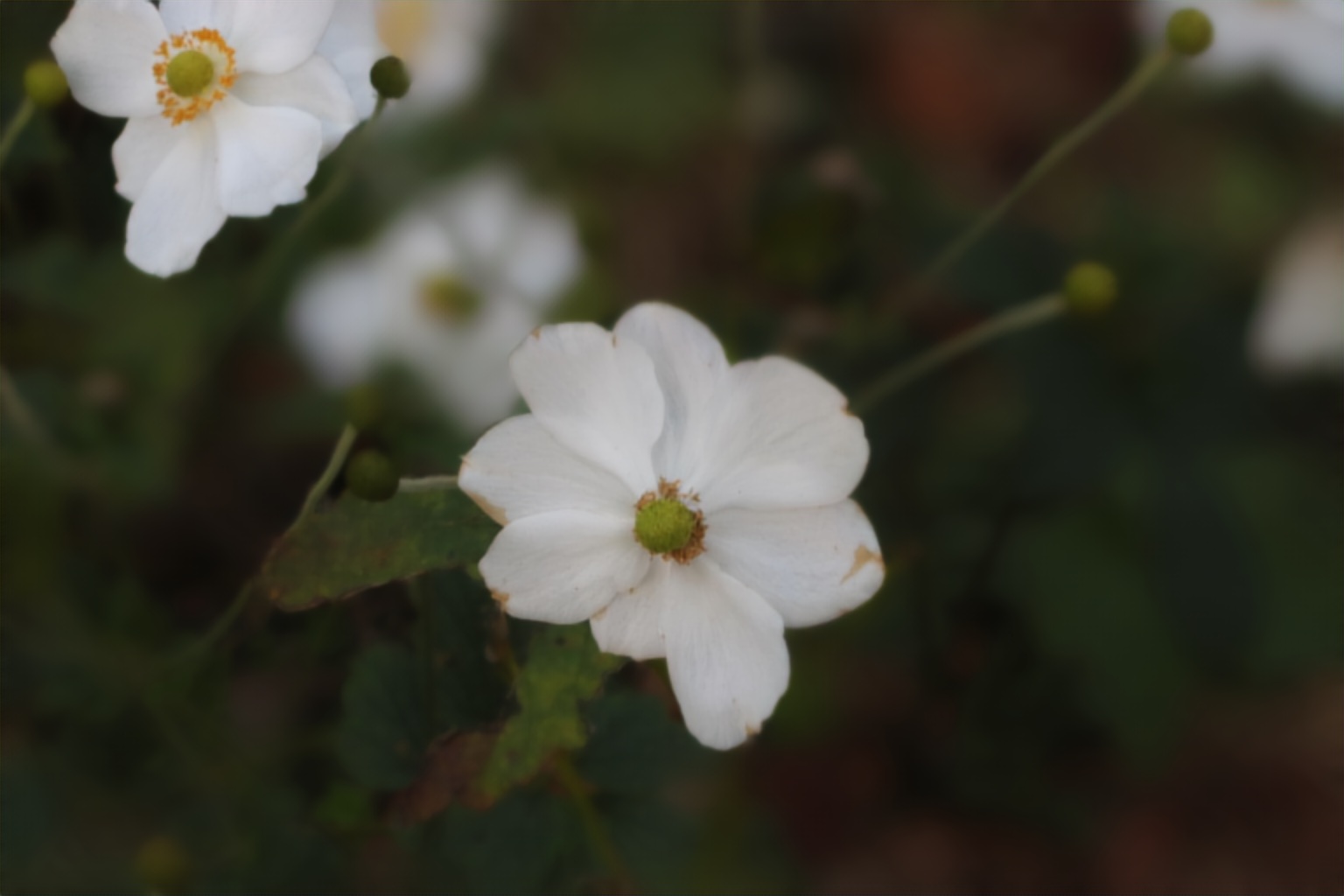}
    \end{subfigure}
    \begin{subfigure}{.24\textwidth}
    \centering
    \includegraphics[width = \linewidth]{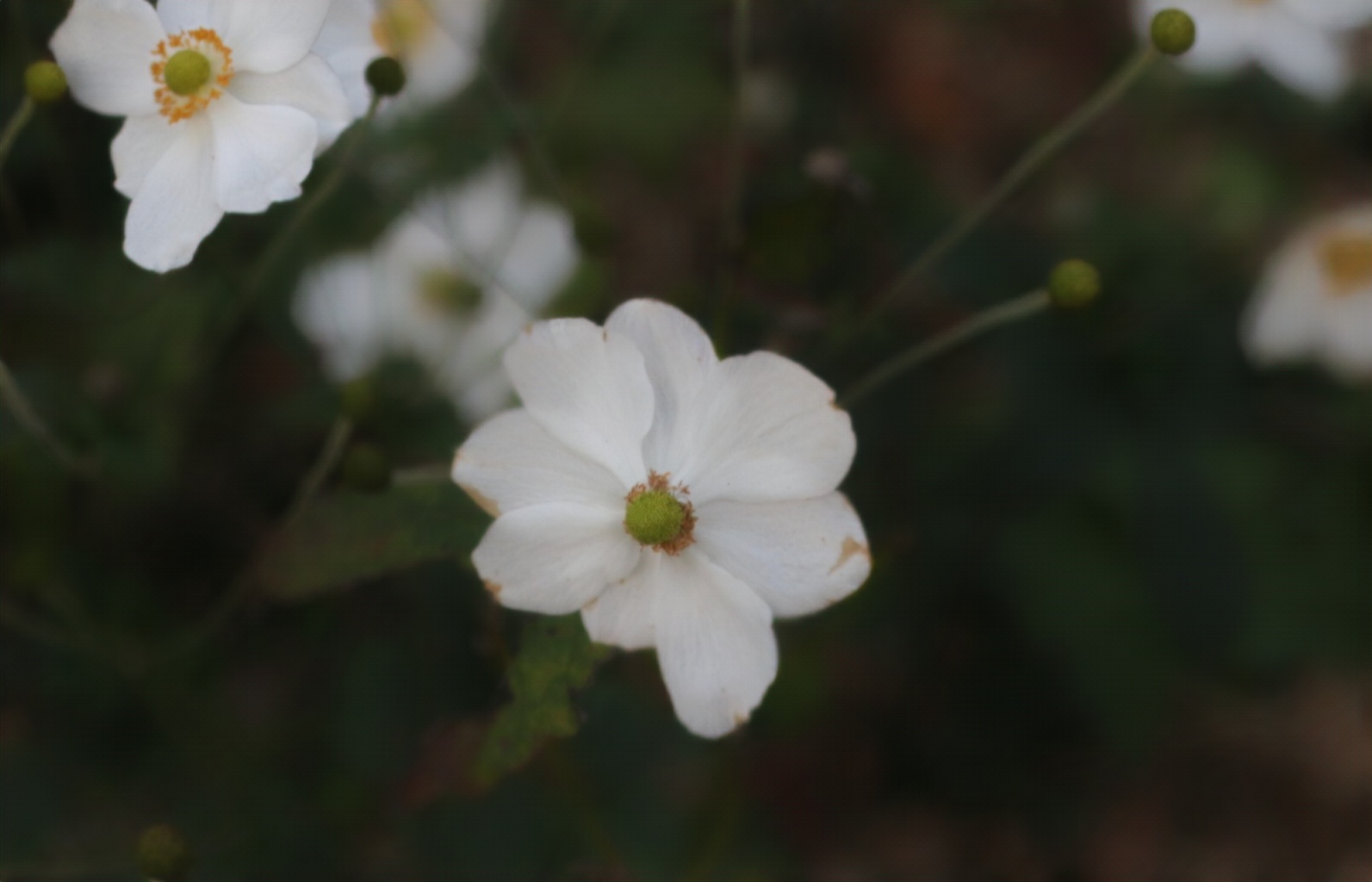}
    \end{subfigure}
    \begin{subfigure}{.24\textwidth}
    \centering
    \includegraphics[width = \linewidth]{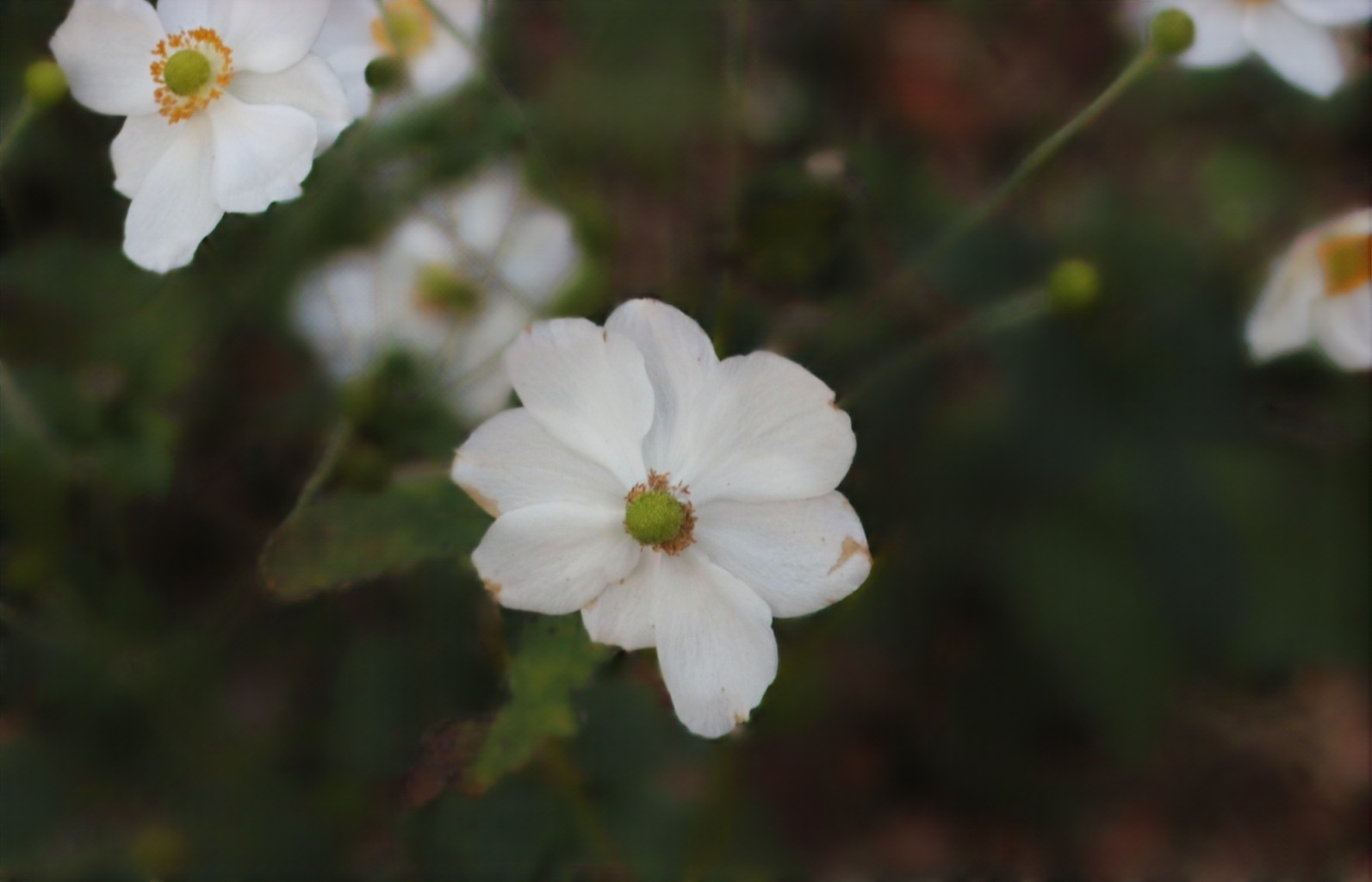}
    \end{subfigure} \\
    \begin{subfigure}{.24\textwidth}
    \centering
    \includegraphics[width = \linewidth]{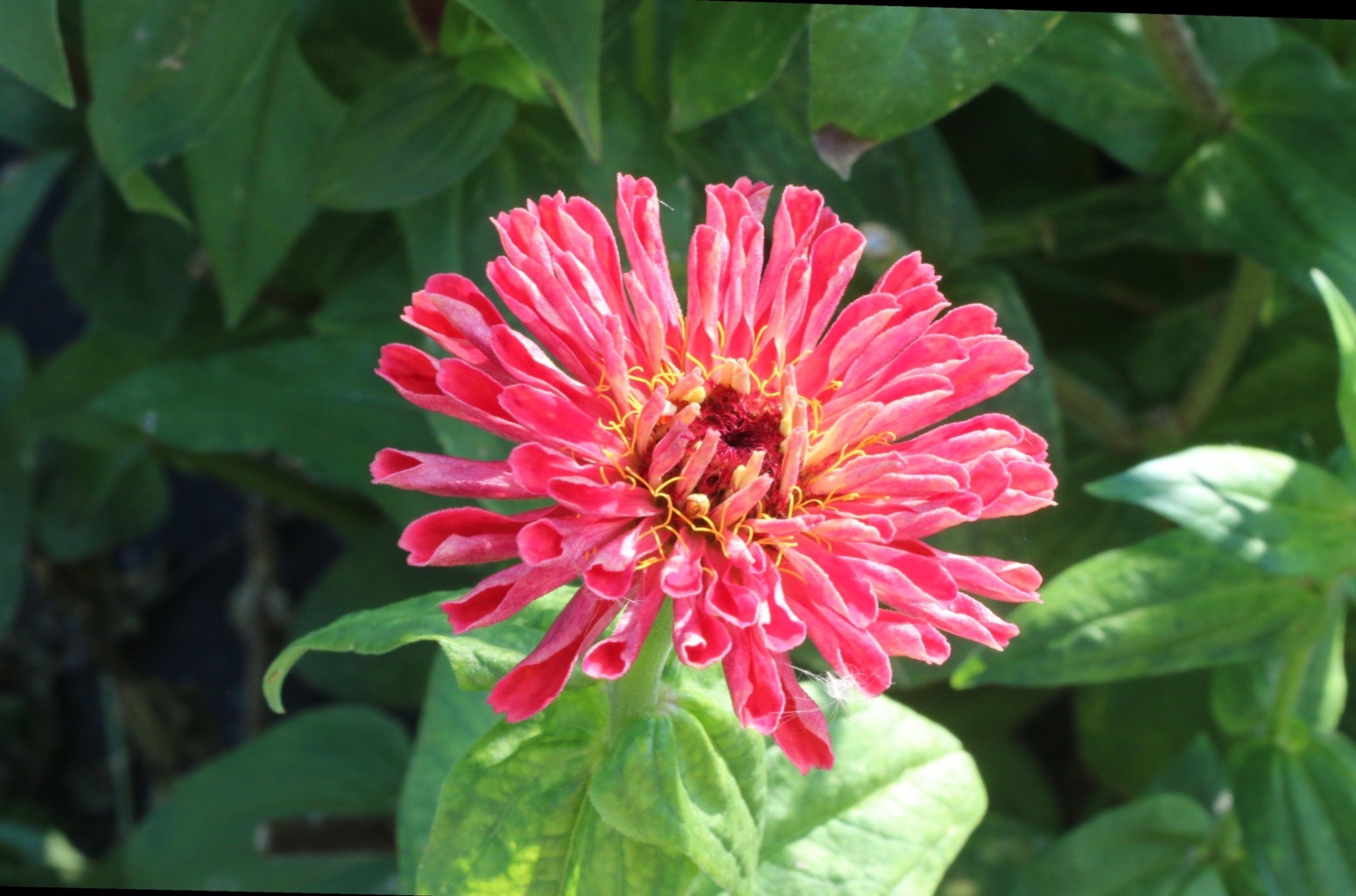}
    \end{subfigure}
    \begin{subfigure}{.24\textwidth}
    \centering
    \includegraphics[width = \linewidth]{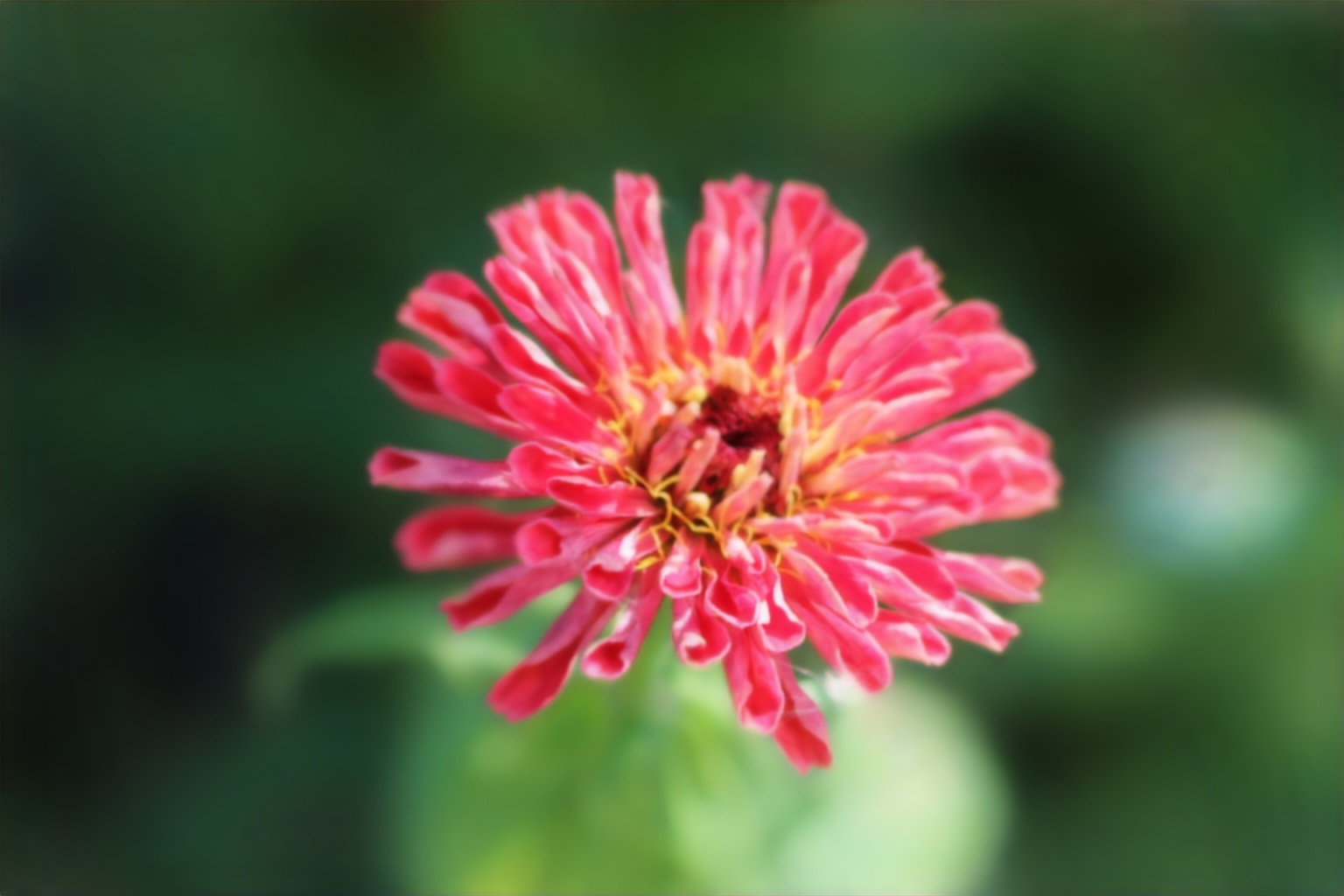}
    \end{subfigure}
    \begin{subfigure}{.24\textwidth}
    \centering
    \includegraphics[width = \linewidth]{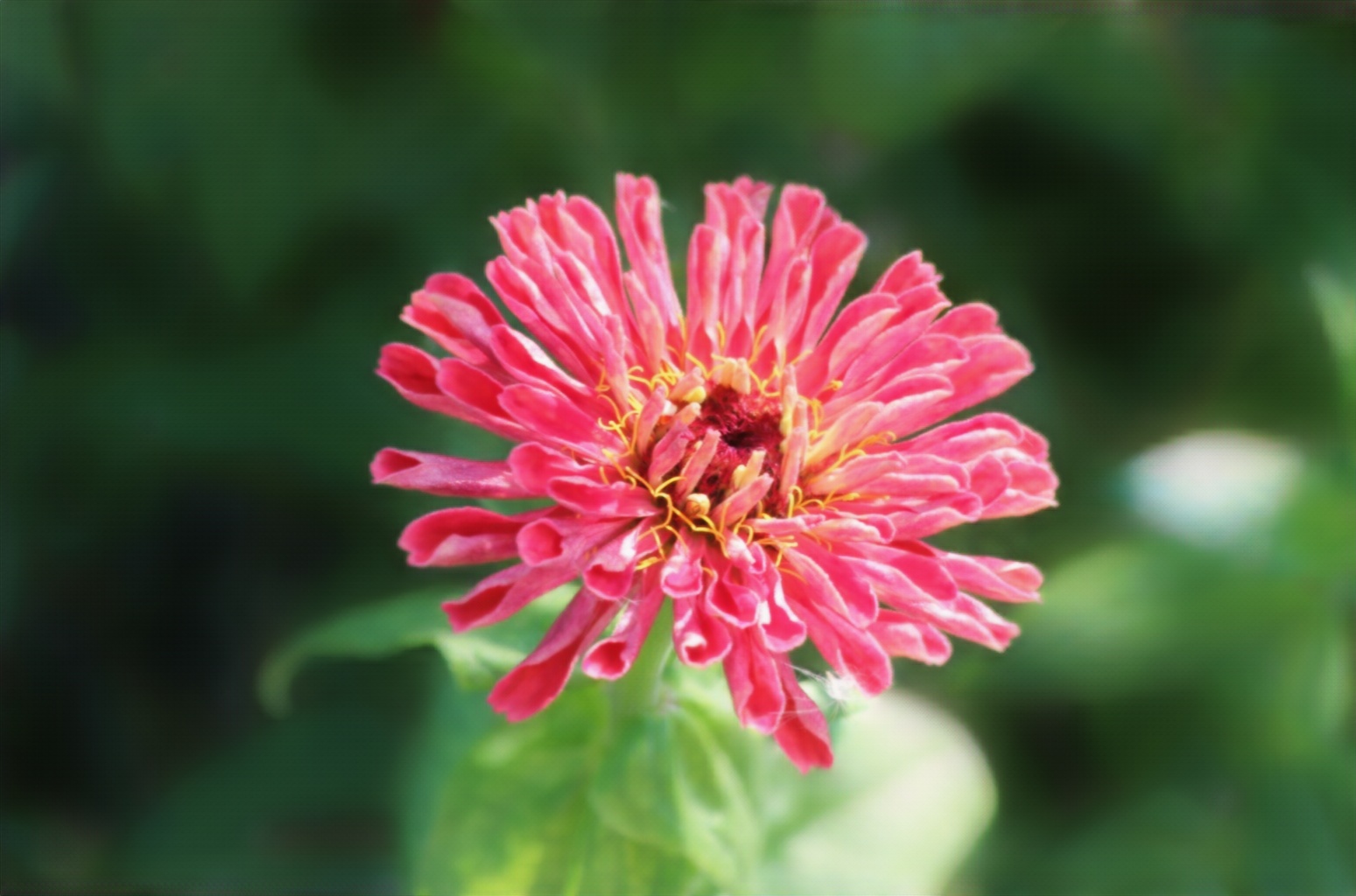}
    \end{subfigure}
    \begin{subfigure}{.24\textwidth}
    \centering
    \includegraphics[width = \linewidth]{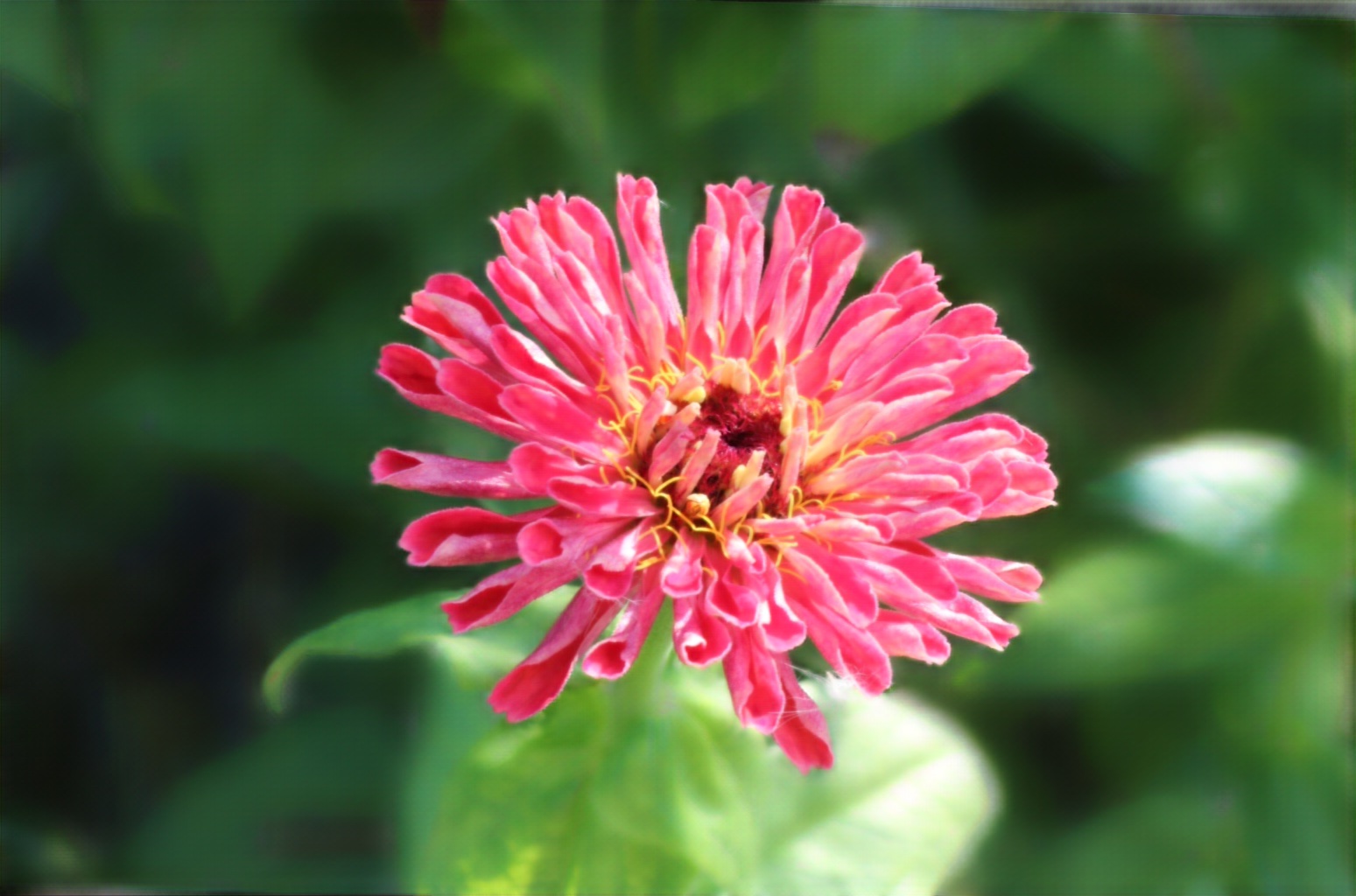}
    \end{subfigure}  \\
    \begin{subfigure}{.24\textwidth}
    \centering
    \includegraphics[width = \linewidth]{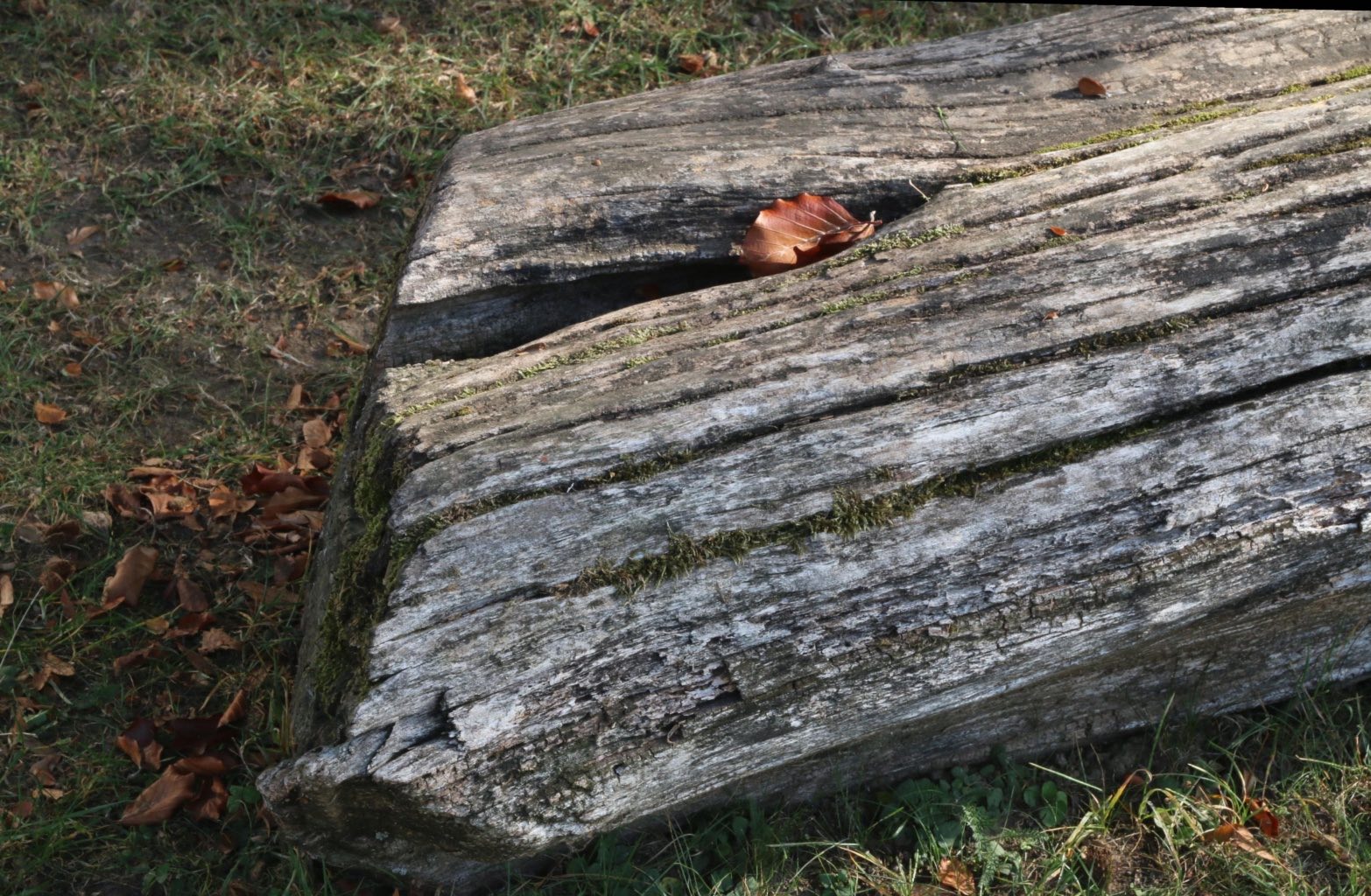}
    \end{subfigure}
    \begin{subfigure}{.24\textwidth}
    \centering
    \includegraphics[width = \linewidth]{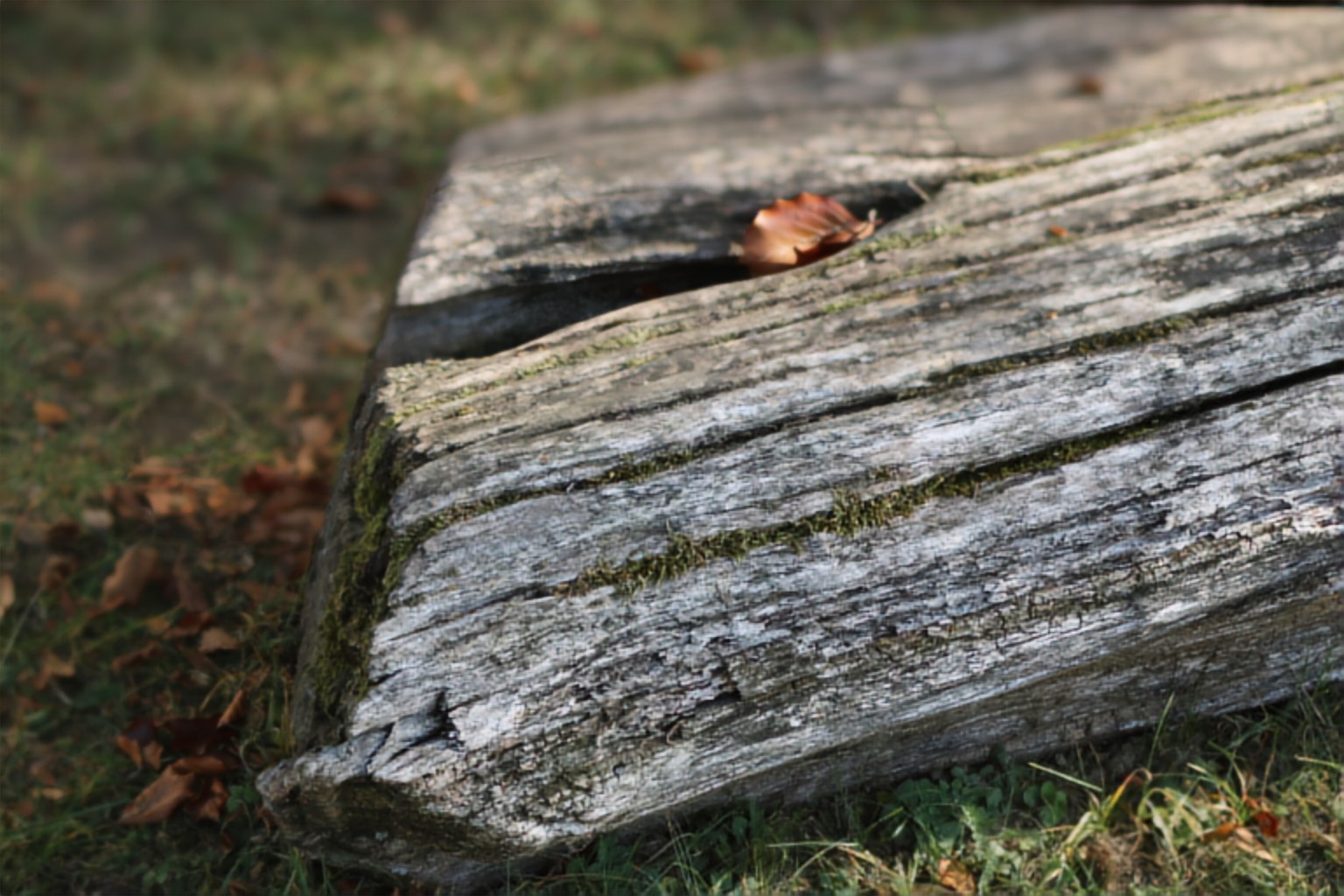}
    \end{subfigure}
    \begin{subfigure}{.24\textwidth}
    \centering
    \includegraphics[width = \linewidth]{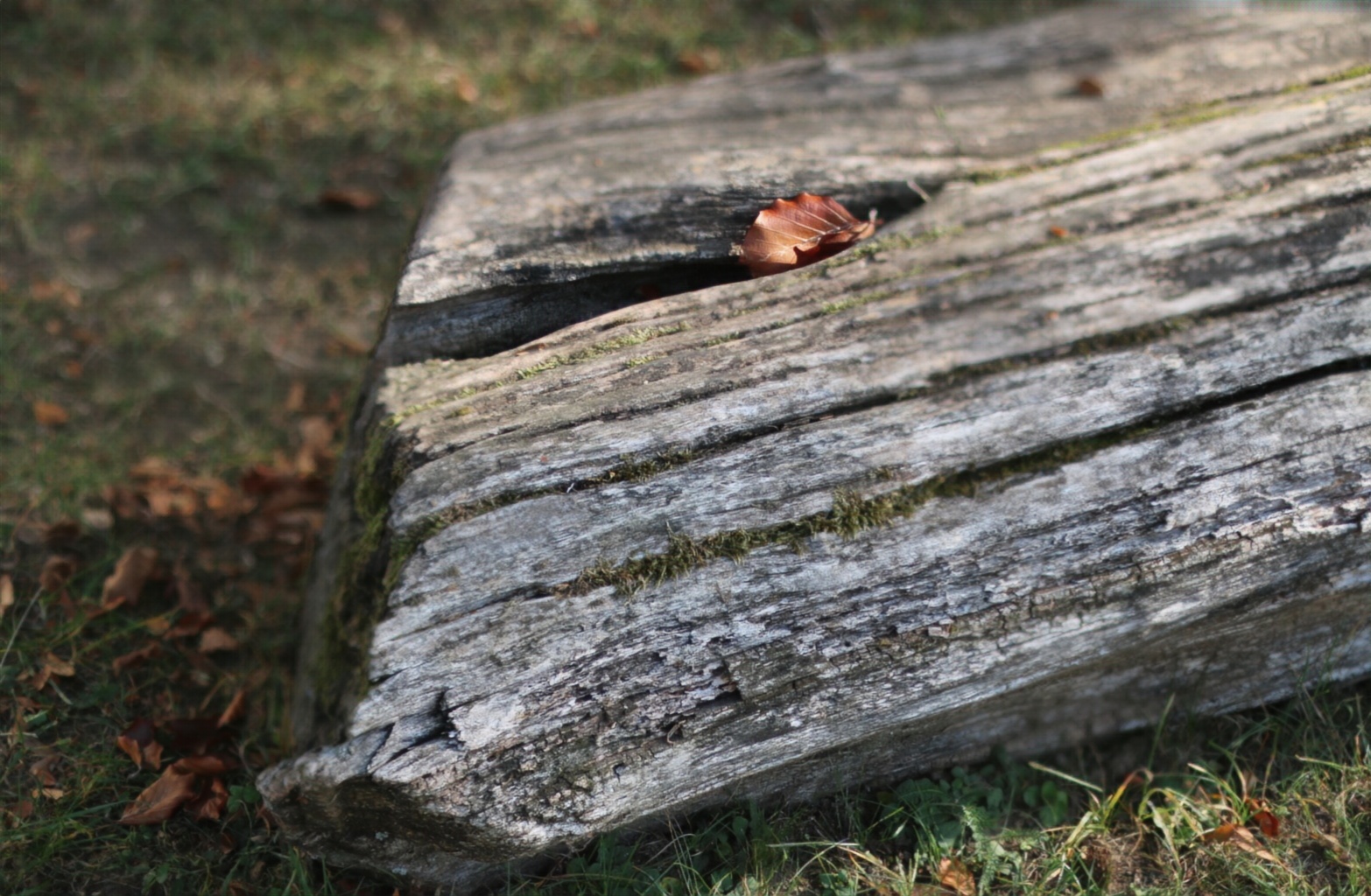}
    \end{subfigure}
    \begin{subfigure}{.24\textwidth}
    \centering
    \includegraphics[width = \linewidth]{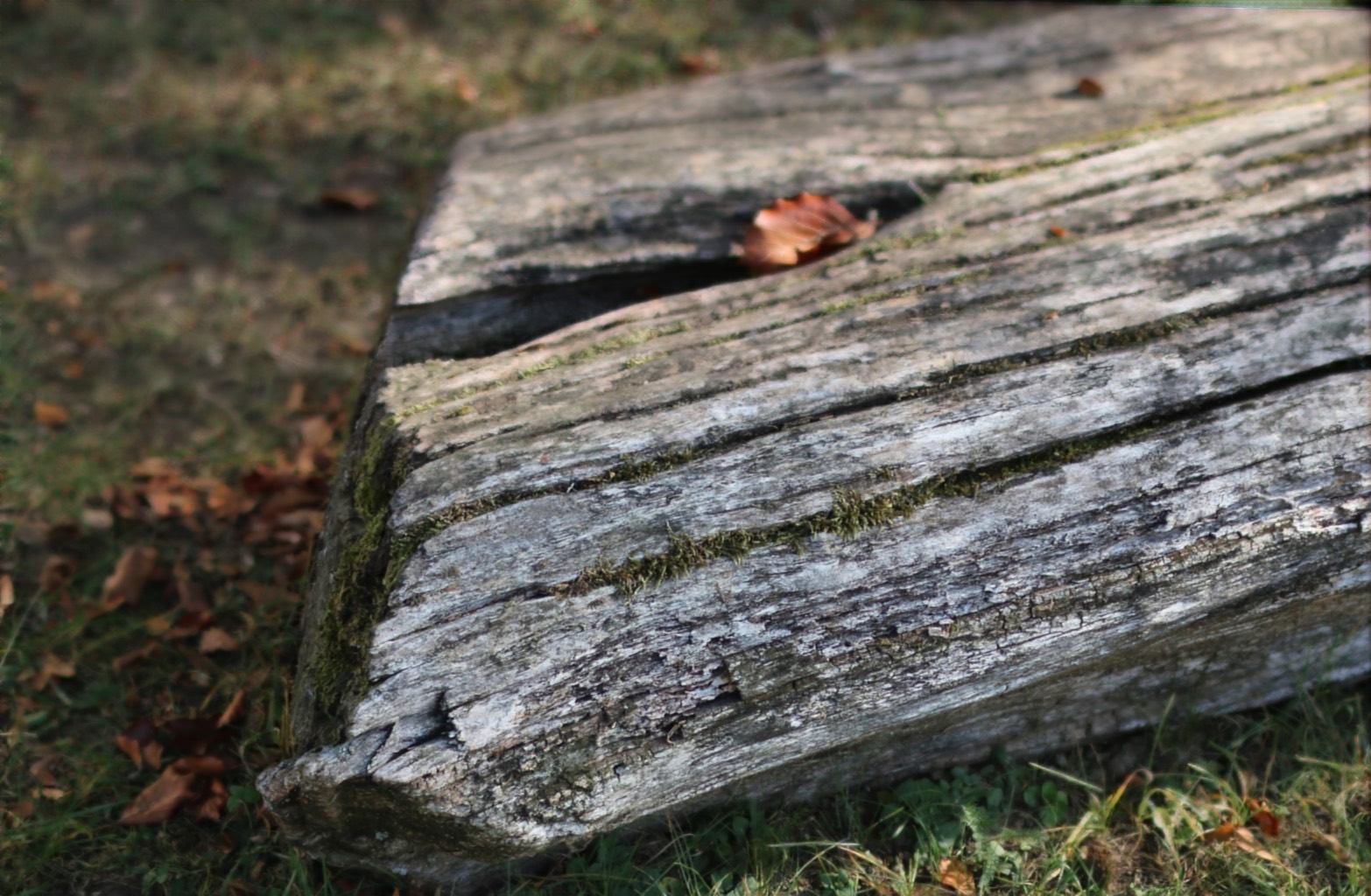}
    \end{subfigure} 
    \end{center}
    \caption{Results of our model on the EBB test dataset. From left to right: input image, results of PyNET, results of BGGAN, our results. Our model is able to attain a strong blur while maintaing a sharp foreground. Best viewed from computer screen.}
    \label{fig:comparison}
\end{figure}

In this section, we provide sample visual results of our model compared to the results of the current state-of-the-art solutions trained specifically for Bokeh rendering \cite{pynet,bggan}. As shown in Figure \ref{fig:comparison}, our model has a blur quality that is similarly natural compared to BGGAN \cite{bggan}, while performing better in quality than PyNET. On the other hand, the edges are sharper and more prominent in our model compared to other models, while the salient regions also remain sharper.

\subsection{Ablation Studies}

As mentioned before, our model consists of various different components and is trained in a multi-stage manner with a combination of different losses and cues. In this section, we analyze the effects of different parts of the model through a comparison of the outputs on the \textit{Val200} validation set. 

\subsubsection{Blurring Cues}
One of the main aspects of our network is the addition of a depth prediction module in DPT to help guide the blur. This is done through a concatenation of the original input image $\textbf{I}$ with the depth map generated by running the input image through a depth estimation module $D$, before running through a backbone. 

Other blur cues, such as a binary saliency mask or a defocus map can also be used, however, to help guide the blur. We thus tried concatenating various different alternatives to the depth map we generated using the DPT module, including a binary saliency mask generated through the TRACER algorithm \cite{TRACER} and a defocus map generated by DMENet \cite{DMENet}.  

As we are only trying to compare the concatenated input map (eg. depth map, saliency map, defocus map) as a blurring clue and not a saliency mask for sharpening during the Bokeh Loss (see \ref{bokehloss_ablation}), we only compare the result of concatenating the input channel and running the NAFNET/GAN setup without any bokeh losses. 

\begin{table}[h!]
\begin{center}
\begin{tabular}{| P{3cm} | P{3cm} | P{3cm} | P{3cm} |}
 \hline
 Blur Cue & PSNR & SSIM & MOS  \\
 \hline
 Depth Map (DPT) & \textbf{23.70} & \textbf{0.867} & \textbf{4.01} \\ 
 \hline
 Saliency Mask (TRACER) & 23.26 & 0.863 & 3.72\\ \hline
 Defocus Map (DMENet) & 22.00 & 0.800 & 3.30\\ 
 \hline
\end{tabular}
\end{center}
\caption{Quantitative and qualitative results of GAN-based training with various different blur cues on the Val200 dataset.}
\label{table: blurcue_ablation}
\end{table}
As shown in Table \ref{table: blurcue_ablation}, we found that the depth map performed the best in terms of blur cues across all metrics. This was largely due to the quality of the depth map that was geenrated, compared to the saliency and defocus maps that were generated through the TRACER \cite{TRACER} and DMENet \cite{DMENet} algorithms respectively. We thus chose the depth map as our main blurring cue.

In particular, we found that the bokeh images generated using the depth map had not only the best blur quality, but also the most natural foreground boundaries. This motivated us to use the depth map for the bokeh loss as well.

\subsubsection{Bokeh Loss}\label{bokehloss_ablation}
To test the effectiveness of the Bokeh losses, we ran an ablation study comparing results of the model with and without them. All models here use the depth map as a blur cue. The second model uses the depth map also as a greyscale mask for the bokeh loss.  

\begin{table}[h]
\begin{center}
\begin{tabular}{| P{3cm} | P{3cm} | P{3cm} | P{3cm} |}
 \hline
 Model & PSNR & SSIM & MOS \\
 \hline
 NAFNet8 (GAN/No Bokeh Loss) & \textbf{23.70} & 0.867 & 4.01 \\ 
 \hline
 NAFNet8 (GAN/Bokeh Loss during pretraining) & 23.475 & \textbf{0.868} & \textbf{4.23}\\
 \hline
\end{tabular}
\end{center}
\caption{Results of different training strategies for the Bokeh Loss}
\label{table: bokehloss}
\end{table}

As shown in table \ref{table: bokehloss}, the addition of the Bokeh loss provides a substantial increase in the MOS, with the edges being noticeably sharper and the background smoother.

\subsubsection{Model Backbone}\label{model_backbone_ablation}

To test the effectiveness of our model backbone, we tested the effect of switching the 8-layer NAFNet backbone with two cascading U-Nets (GlassNet) similar to in \cite{bggan}. Both models were trained with the full pipeline, using the Bokeh losses in the pretraining stage (as described in section \ref{method}) and the dual-scale GAN in the refinement stage. 

\begin{table}[h!]
\begin{center}
\begin{tabular}{| P{2.4cm} | P{2.4cm} | P{2.4cm} | P{2.4cm} | P{2.4cm} |}
 \hline
 Backbone & No. of Parameters & PSNR & SSIM & MOS \\
 \hline
 GlassNet & 10.353M & 22.956 & \textbf{0.875} & \textbf{4.25}\\ 
 \hline
 8-Layer NAFNet & \textbf{1.047M} & \textbf{23.475} & 0.868 & 4.23\\
 \hline
\end{tabular}
\end{center}
\caption{Results of different backbones for the pipeline}
\label{table: backbone}
\end{table}

We found that the 8-layer NAFNet performed almost identically similar to the GlassNet proposed in \cite{bggan} despite having only around a tenth of the parameters. This justified our choice of switching the model backbone for the NAFNet.

\subsubsection{NAFNet Width}\label{Nafnet_size_ablation}
As discussed in section \ref{NAFNet}, the NAFNet contains a variable number of width and number of enclosing/decoding/middle blocks. The main two factors in determining model complexity are the width of the NAFBlocks and the number of NAFBlocks. As discussed in \cite{nafnet}, an increase in the number of blocks has shown to not increase latency greatly while greatly improving model performance on the original image restoration task. On the other hand, due to memory constraints of mobile devices, the bottleneck in the model prediction is instead the width of the NAFBlocks, due to the amount of processing needed for wider blocks.  

We thus chose to reduce the width of the individual NAFBlocks from the original NAFBlock width of $32$. This reduction of width, however, must be balanced with our desire to process the input images at multiple (global and local) scales. Based on these considerations, we ran an ablation study to find a suitable balance between model width and computational complexity. 

\begin{table}[h!]
\begin{center}
\begin{tabular}{| P{2.4cm} | P{2.4cm} | P{2.4cm} | P{2.4cm} | P{2.4cm} |}
 \hline
 NAFBlock Width & No. of Parameters & PSNR & SSIM & MOS \\
 \hline
 4 & \textbf{277K} & \textbf{23.603} & 0.850& 3.57\\
 \hline
 8 & 1.047M & 23.475 & \textbf{0.868} & 4.23 \\ 
 \hline
 12 & 2.311M & 23.298 & 0.866 & \textbf{4.31}\\ \hline
\end{tabular}
\end{center}
\caption{Quantitative and qualitative results of $36$-block NAFNet with various different widths}
\label{table: nafnet_width}
\end{table}

Due to model size considerations, we found that the backbone needed to have a width of $12$ or lower to fit on a mobile device. As shown in table \ref{table: nafnet_width}, there is not a significant difference in MOS for widths above $8$ layers, while the parameter count increases greatly, unlike in the shift from a width of $4$ to $8$ (where there is a big increase in MOS). We thus chose a final NAFBlock width of $8$. 

\section{Conclusions}

We have presented a new method for monocular bokeh synthesis to generate natural shallow depth-of-field images. By using a monocular depth estimate as both a blur cue and a greyscale saliency mask for three new ``bokeh-specific'' loss functions, our model was able to produce images with sharp foreground edges and a smooth background. Coupled with a multi-receptive field conditional adversarial network \cite{Pix2pix} as part of a dual-stage training process, our model was able to achieve a sharp, yet natural bokeh render. Extensive ablations are then presented to validate the effect of each of the individual modules. In the future, we can explore adding radiance modules to help render the bokeh effect in a more physics-based manner, along with using a smaller depth-prediction module to decrease the inference time on TFLite.   

\appendix

\section{Pytorch to TFLite Conversion}\label{appendix1}
Our model was converted from Pytorch to TFLite through a two-part process. First, we converted the trained NAFNet model using the Pytorch $\rightarrow$ Onnx $\rightarrow$ Tensorflow pipeline, and then combined a pretrained DPT net (which was pre-written in Tensorflow) in tensorflow with our saved Tensorflow NAFNet module. We then ran a conversion from Tensorflow $\rightarrow$ TFLite by first saving the resulting Tensorflow Module as a frozen graph and then exporting it to TFLite using the version 2 converter. 

\section*{Acknowledgements} We would like to thank the team at Sensebrain Technology for their helpful suggestions. We did not receive external funding or additional revenues for this project.
%
%
\bibliographystyle{splncs04}
\bibliography{egbib}
\end{document}